\documentclass{article}

\usepackage{tikz}
\usetikzlibrary{positioning, arrows.meta}

\usepackage[preprint]{neurips_2024}
\usepackage[utf8]{inputenc} 
\usepackage[T1]{fontenc}    
\usepackage{hyperref}       
\usepackage{url}            
\usepackage{booktabs}       
\usepackage{makecell}
\usepackage{siunitx}

\usepackage{amsfonts}       
\usepackage{nicefrac}       
\usepackage{microtype}      
\usepackage{xcolor}         
\usepackage{enumitem}
\usepackage{amsmath} 

\usepackage{graphicx}
\usepackage{subcaption}

\usepackage[textwidth=2.5cm]{todonotes}

\title{
Duo-LLM: A Framework for Studying \\
Adaptive Computation in Large Language Models}

\author{%
  Keivan Alizadeh  \\
  \And
  Iman Mirzadeh \thanks{Major contribution. Correspondence to \{kalizadehvahid, imirzadeh, farajtabar\}@apple.com.} \\
  \And
  Hooman Shahrokhi\thanks{Work done during an internship at Apple.} \\
  \And
  Dmitry Belenko \\
  \And
  Frank Sun \\
  \And
  Minsik Cho \\
  \And 
  Mohammad Sekhavat \\
  \And
  Moin Nabi \\
  \And
  Mehrdad Farajtabar 
  \AND  Apple
}

\begin{document}

\maketitle

\begin{abstract}

Large Language Models (LLMs) typically generate outputs token by token using a fixed compute budget, leading to inefficient resource utilization. To address this shortcoming, recent advancements in mixture of expert (MoE) models, speculative decoding, and early exit strategies leverage the insight that computational demands can vary significantly based on the complexity and nature of the input. However, identifying optimal routing patterns for dynamic execution remains an open challenge, limiting the full potential of these adaptive methods. To address this need, we study adaptive computation in LLMs more systematically.
We propose a novel framework that integrates smaller auxiliary modules within each Feed-Forward Network layer of the LLM. This design enables dynamic routing of tokens based on task complexity: tokens can be processed by either the small or big modules at each layer, or even bypass certain layers entirely. This allows us to introduce a novel notion of a token's difficulty, defined by its potential to benefit from additional computational resources. Importantly, by employing oracles to identify optimal patterns of adaptive computations, we gain valuable insights into the internal workings of LLMs and the routing processes in a simplified heterogeneous MoE setup. We show that trained routers operate differently from oracles and often yield suboptimal solutions. Notably, activating a large module in just one layer outperforms models that use large modules across all layers, underscoring the gap between practical implementations of routing in MoE models and theoretical optima for adaptive computation.

\end{abstract}

\begin{figure*}[ht]
  \centering
  \begin{subfigure}[t]{0.4\linewidth}
    \includegraphics[width=\linewidth]{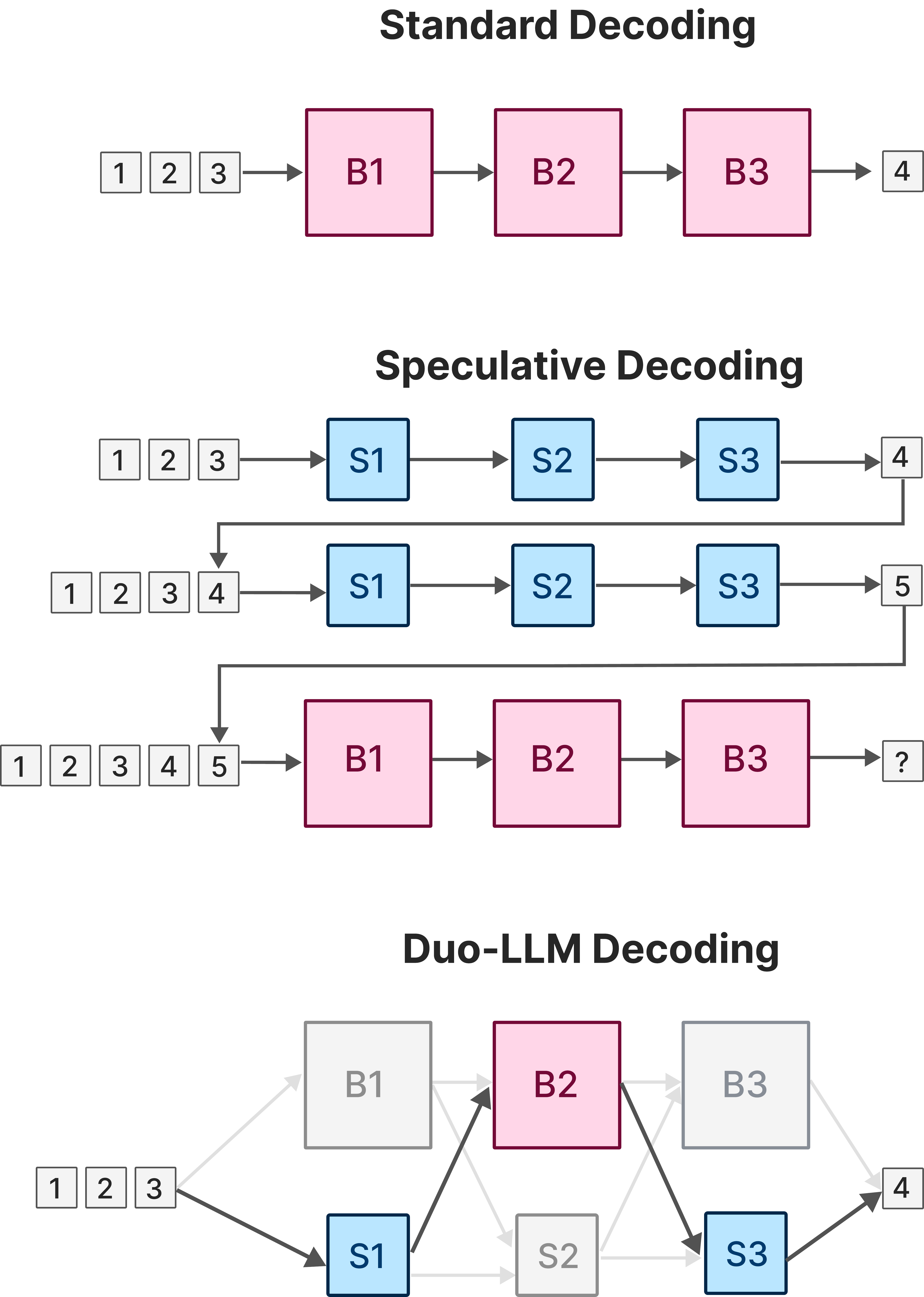}
    \caption{Different decoding methods}
    \label{fig:duo-llm-framework}
\end{subfigure}
\hfill
\begin{subfigure}[t]{0.51\linewidth}
    \includegraphics[width=\linewidth]{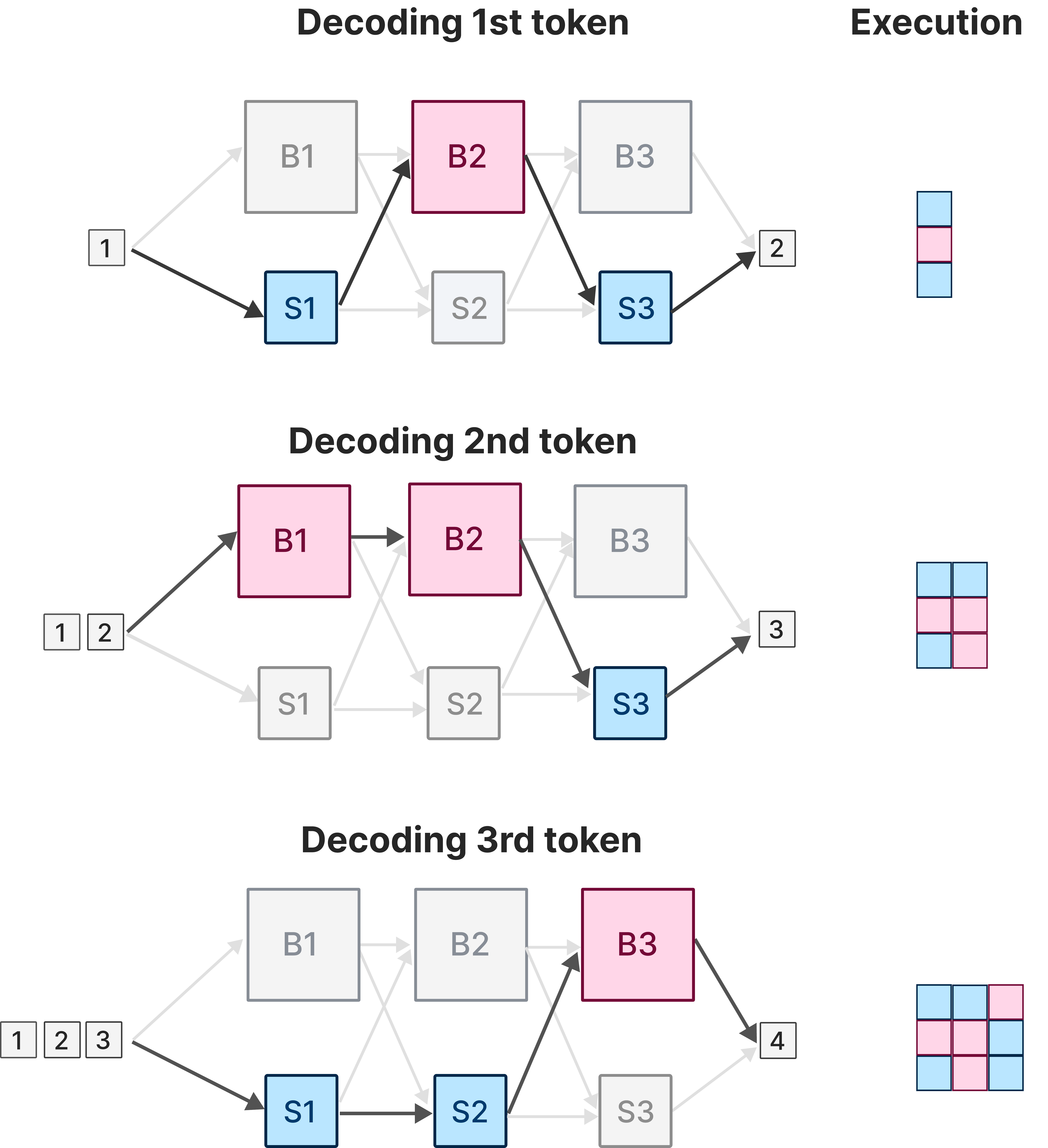}
    \caption{Duo-LLM decoding in action}
    \label{fig:duo-llm-decoding}
\end{subfigure}

\caption{Duo-LLM Framework: (a) Duo-LLM adds small auxiliary modules to the bigger modules to be used during decoding. (b) Oracle or router is used to study the routing in adaptive computation.}
\label{fig:framework}

\end{figure*}

\section{Introduction}
\label{sec:intro}

Large Language Models (LLMs) have emerged as a pivotal technology in natural language processing, offering unprecedented performance across a diverse array of tasks, from text generation to machine translation and question answering~\citep{brown2020language, raffel2020exploring, devlin2019bert}. These models have demonstrated remarkable capabilities in various domains, showcasing their versatility and potential impact on the field \citep{chowdhery2022palm}. However, the fixed computational budget allocated to process each token in these models remains a significant limitation. Regardless of the input’s complexity or the specific task, the same amount of computational power is utilized, leading to inefficiencies in resource utilization. This rigid approach not only imposes high computational costs but also limits the potential for optimizing model performance.

Recent research has begun to challenge this one-size-fits-all paradigm by exploring adaptive computation mechanisms in neural networks. For instance, speculative decoding \citep{leviathan2022fast} generates future tokens using smaller model and verifies them using the bigger model. Mixture of experts (MoE) models selectively activate different subsets of the model’s parameters, reducing the computational burden while maintaining high performance \citep{shazeer2017outrageously}. Similarly, dynamic and hierarchical models have been proposed to enable more efficient processing by adjusting computation based on the input’s complexity \citep{raposo2024mixtureofdepths, pan2024eetuning}. Additionally, the exploration of layer bypassing and modular architectures has shown that it is possible to significantly reduce computational costs while preserving accuracy \citep{lu2024experts, yun2024inferenceoptimal, Lee2023OrchestraLLMEO, Gong2024MixtureofModulesRT, Lin2024MoMaEE, Wu2024RoutingEL}. These approaches are driven by the recognition that different tokens and tasks often require varying levels of computational resources \citep{salehi2023sharcs,cai2024flextron}.

\begin{figure*}[t]
\begin{subfigure}[t]{0.245\linewidth}
    \centering
    \includegraphics[width=\linewidth]{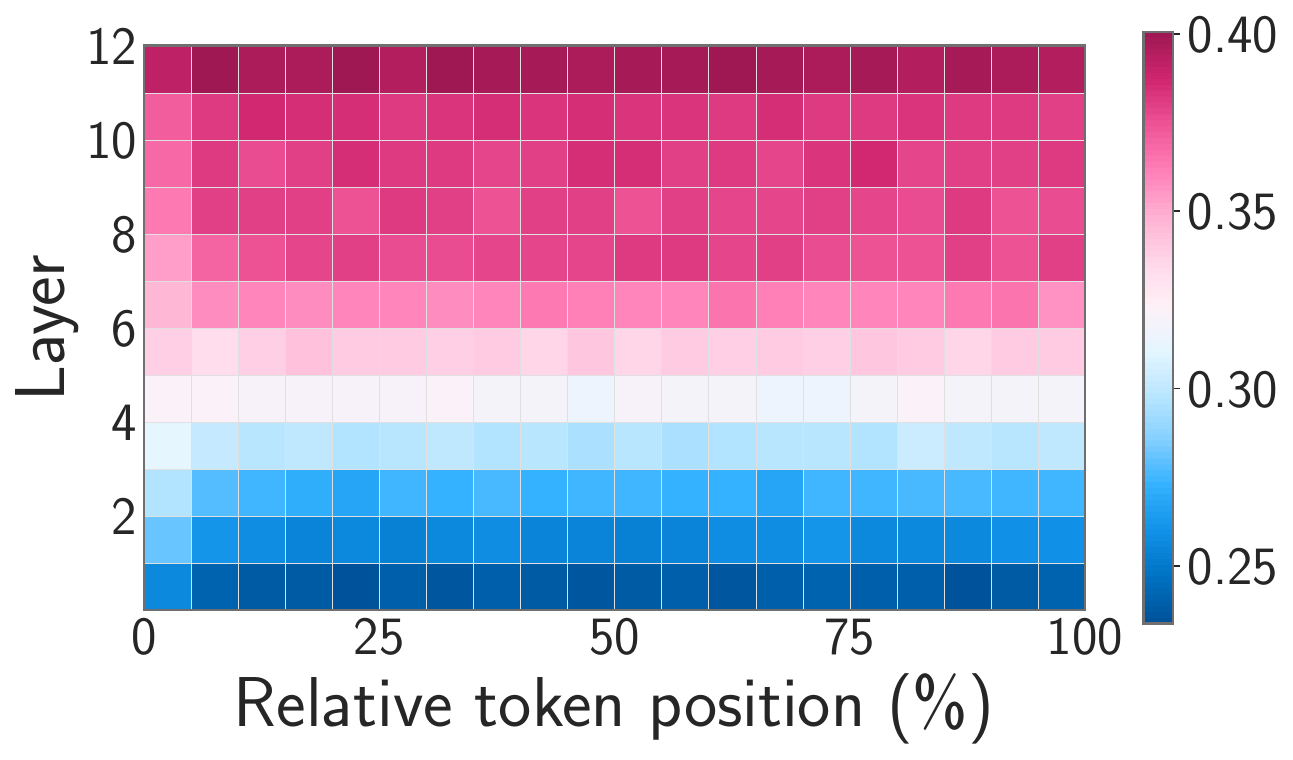}
    \includegraphics[width=\linewidth]{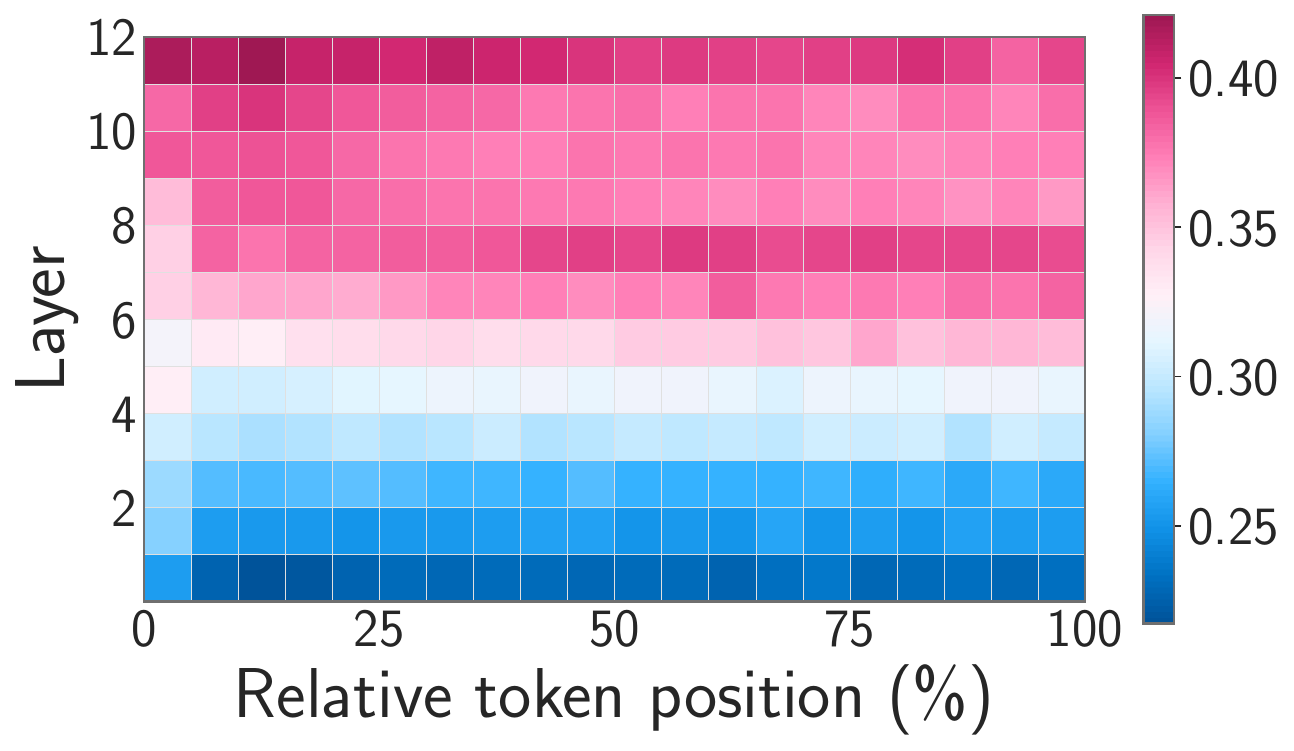}
    \caption{4 big layers}
    \label{fig:fig-combined-33}
\end{subfigure}
\begin{subfigure}[t]{0.245\linewidth}
    \centering
    \includegraphics[width=\linewidth]{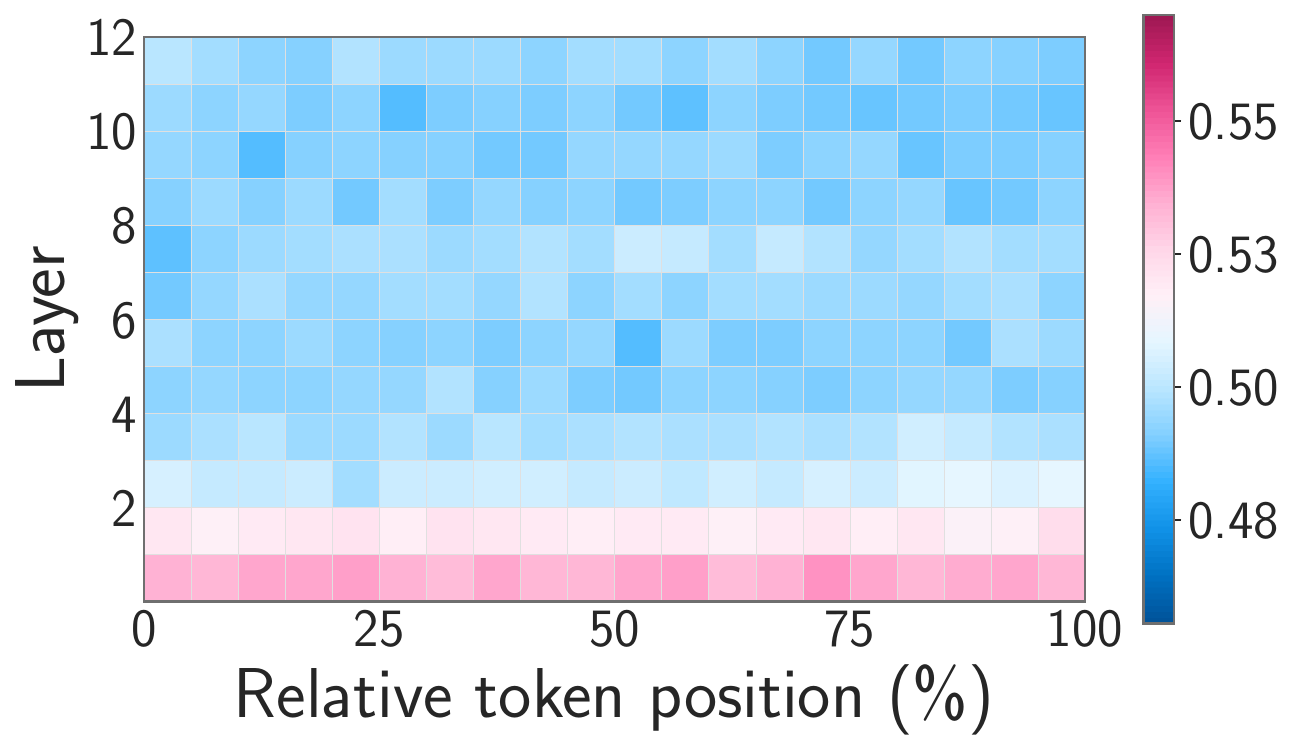}
    \includegraphics[width=\linewidth]{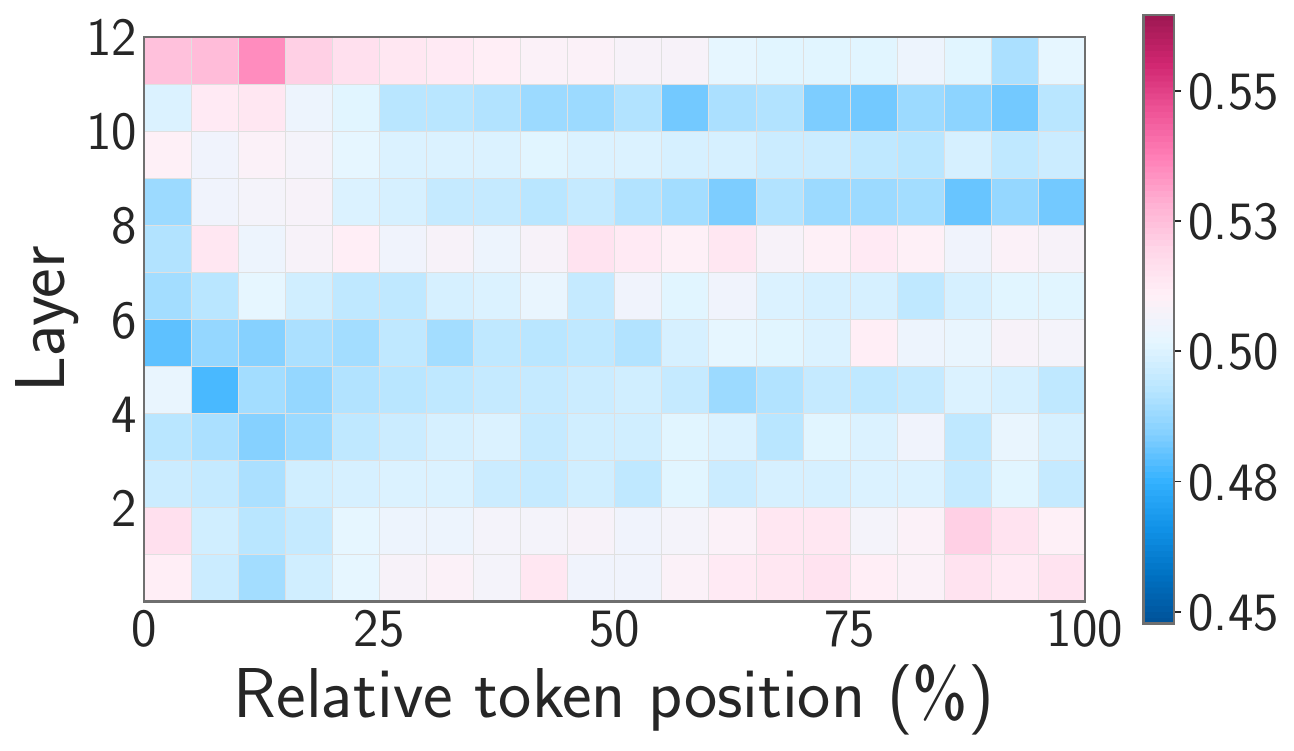}
    \caption{6 big layers}
    \label{fig:fig-combined-55}
\end{subfigure}
\begin{subfigure}[t]{0.244\linewidth}
    \centering
    \includegraphics[width=\linewidth]{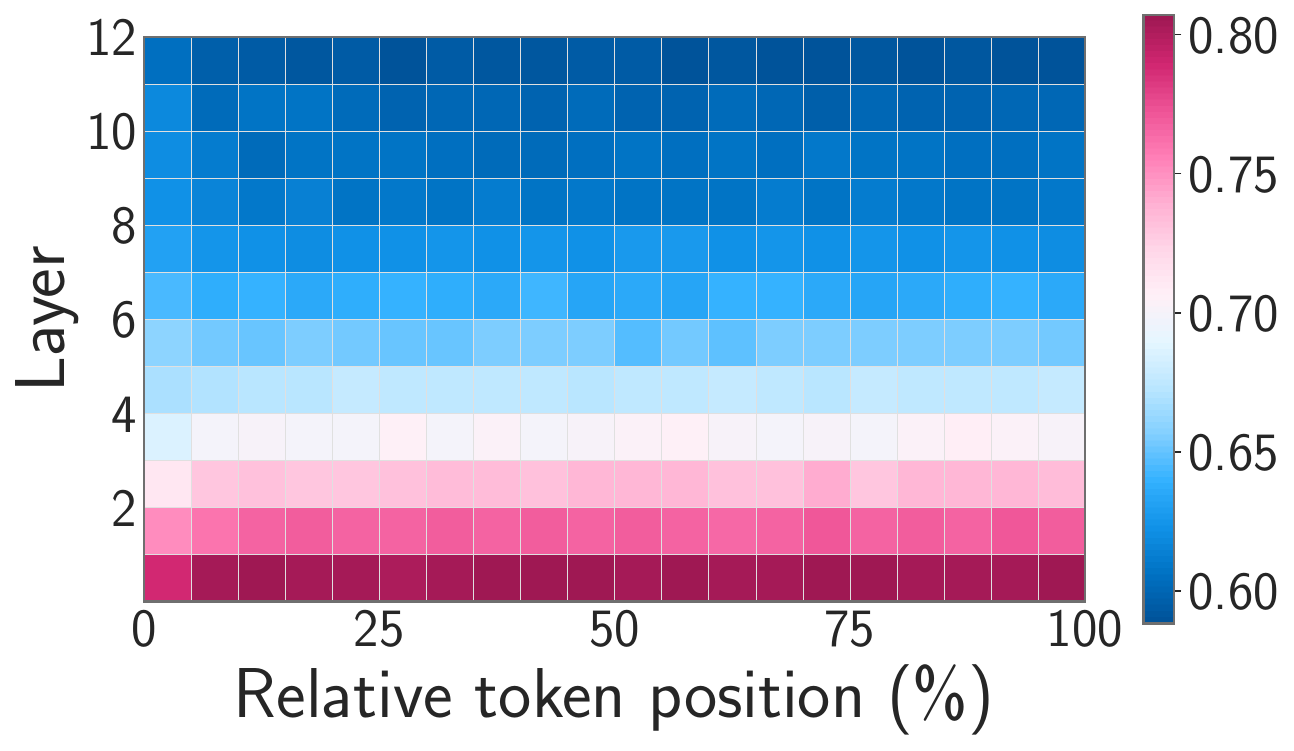}
    \includegraphics[width=\linewidth]{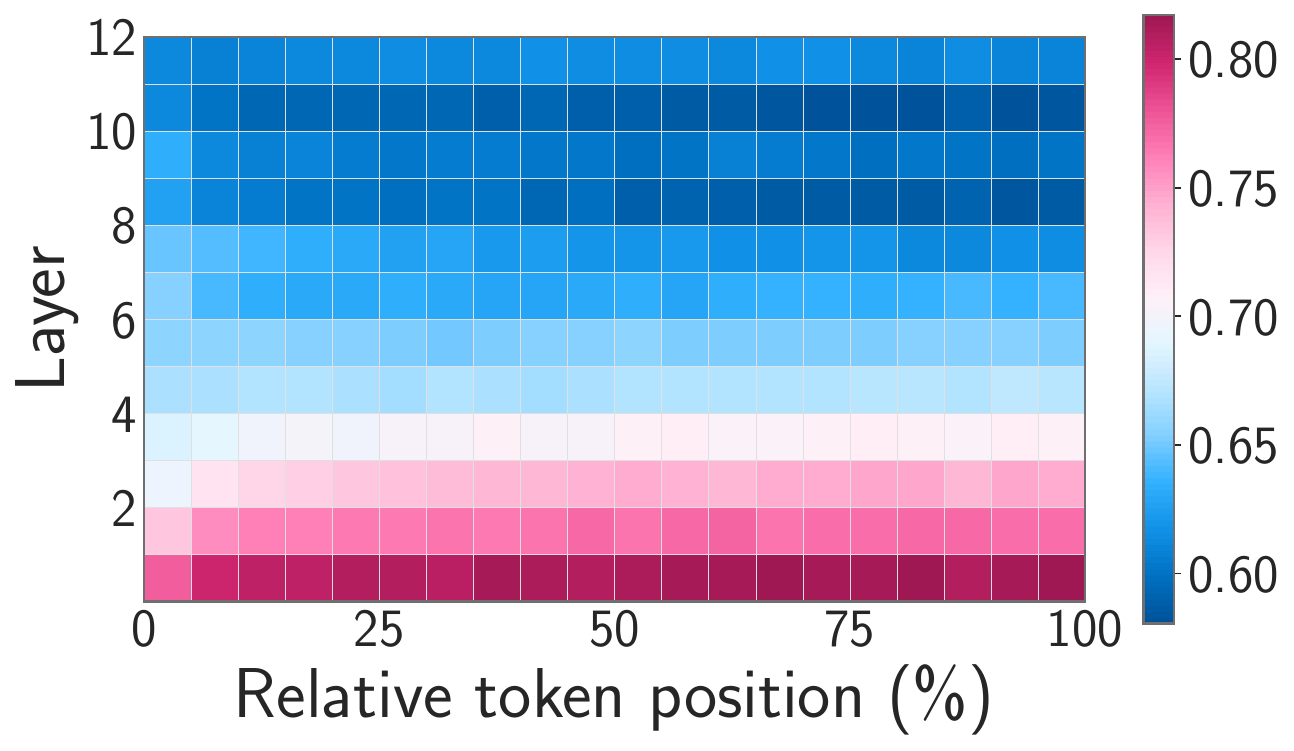}
    \caption{8 big layers}
    \label{fig:fig-combined-77}
\end{subfigure}
\begin{subfigure}[t]{0.25\linewidth}
    \centering
    \includegraphics[width=\linewidth]{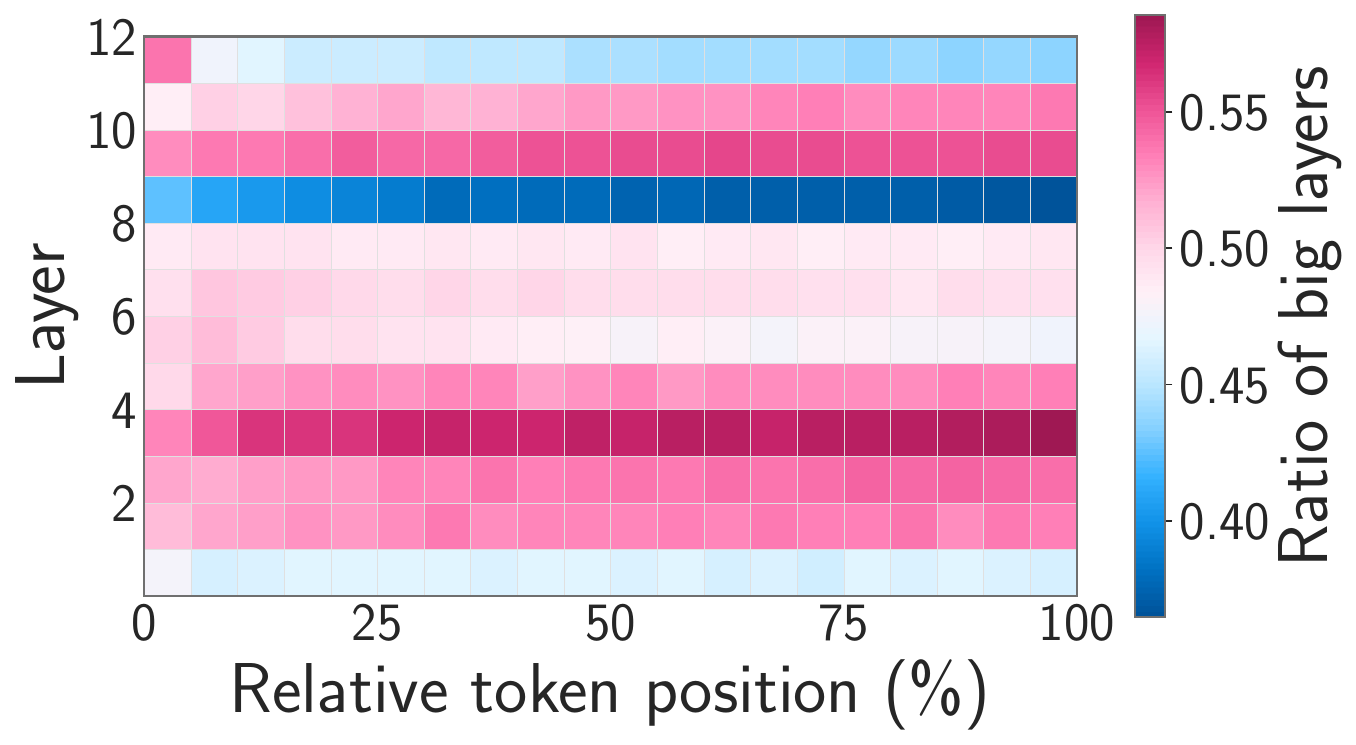}
    \includegraphics[width=\linewidth]{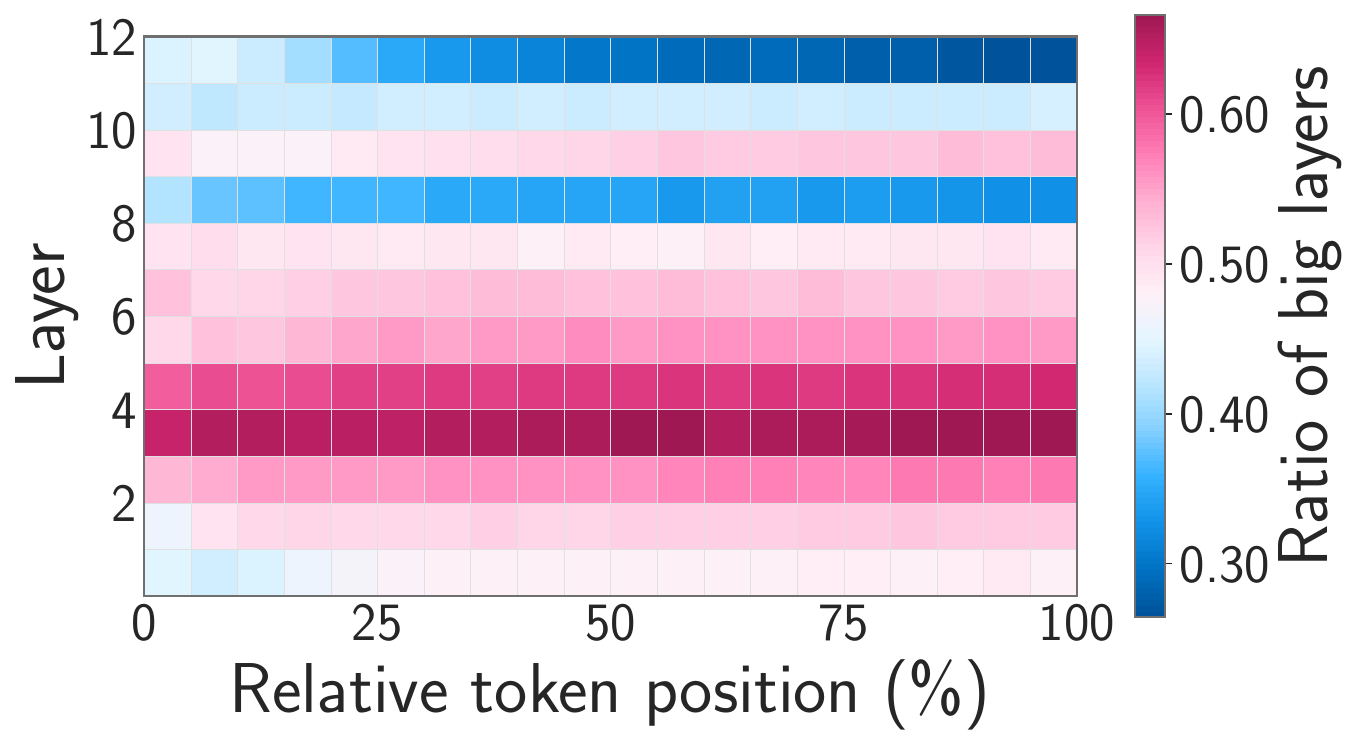}
    \caption{Skip layers}
    \label{fig:fig-combined-skip}
\end{subfigure}
\caption{ 
Oracle routing patterns on the C4 holdout set (top) and the code holdout set (bottom) under different budgets of big layers. (a) With a budget of 4 big layers per token, later layers are chosen more frequently. (b) With a budget of 6 big layers, usage is nearly evenly distributed across all layers. (c) With a budget of 8 big layers, earlier layers are utilized more often. (d) When 6 layers are skipped, some layers are consistently used, but no clear pattern emerges in their ordering.
}
\label{fig:oracle-analysis}
\end{figure*}

\begin{figure*}[t]
\begin{subfigure}[t]{0.245\linewidth}
    \centering
    \includegraphics[width=\linewidth]{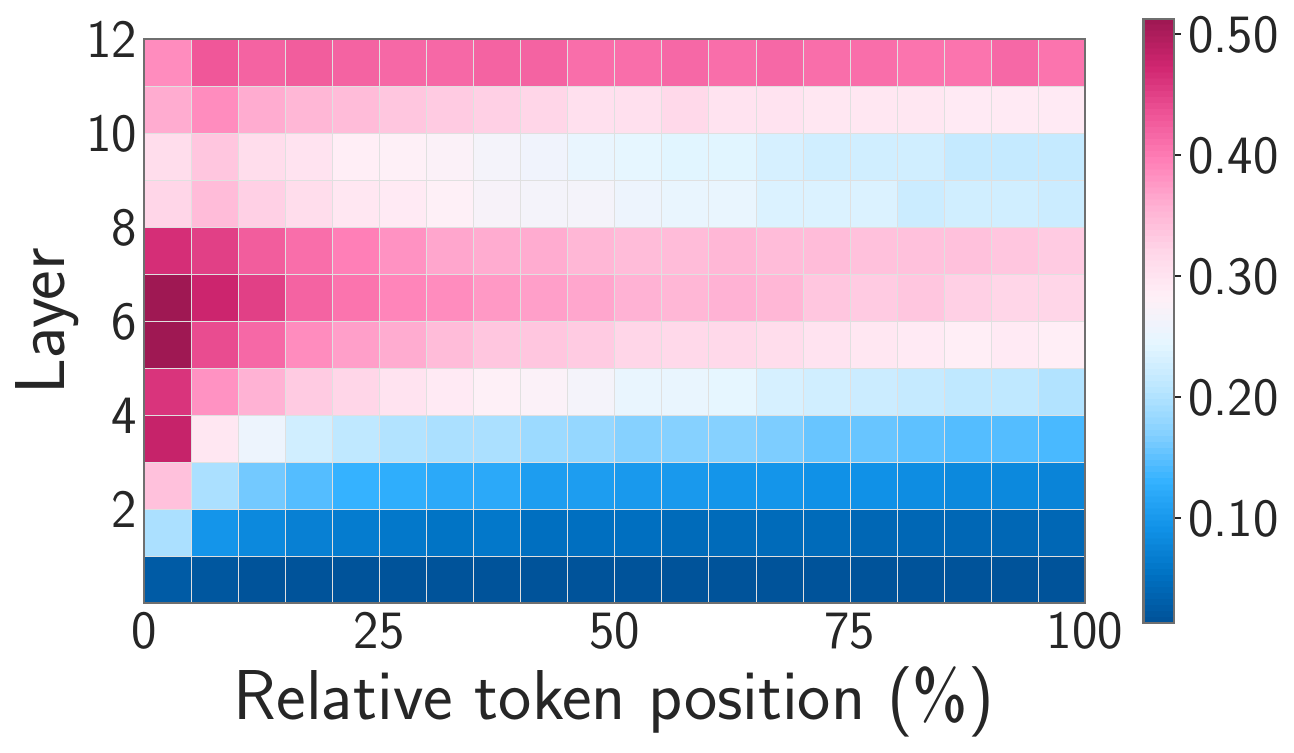}
    \includegraphics[width=\linewidth]{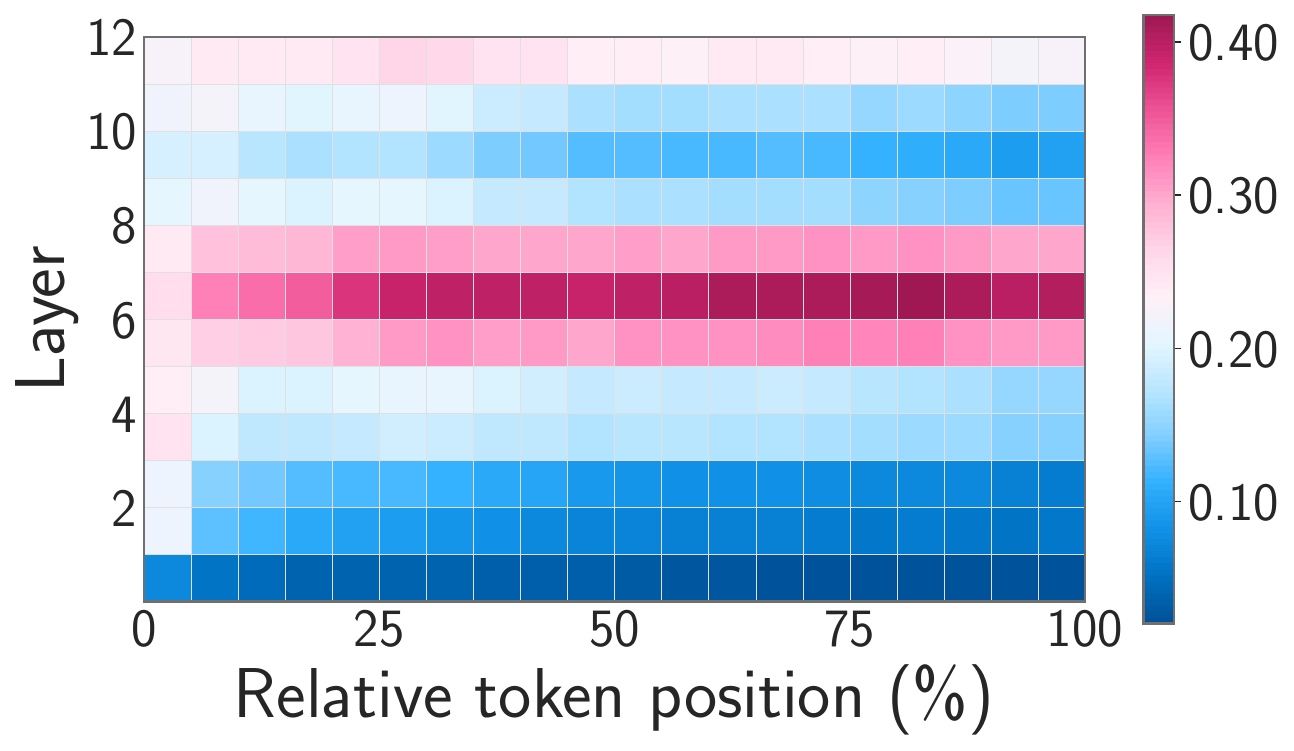}
    \caption{4 big layers}
    \label{fig:fig-combined-33router}
\end{subfigure}
\begin{subfigure}[t]{0.245\linewidth}
    \centering
    \includegraphics[width=\linewidth]{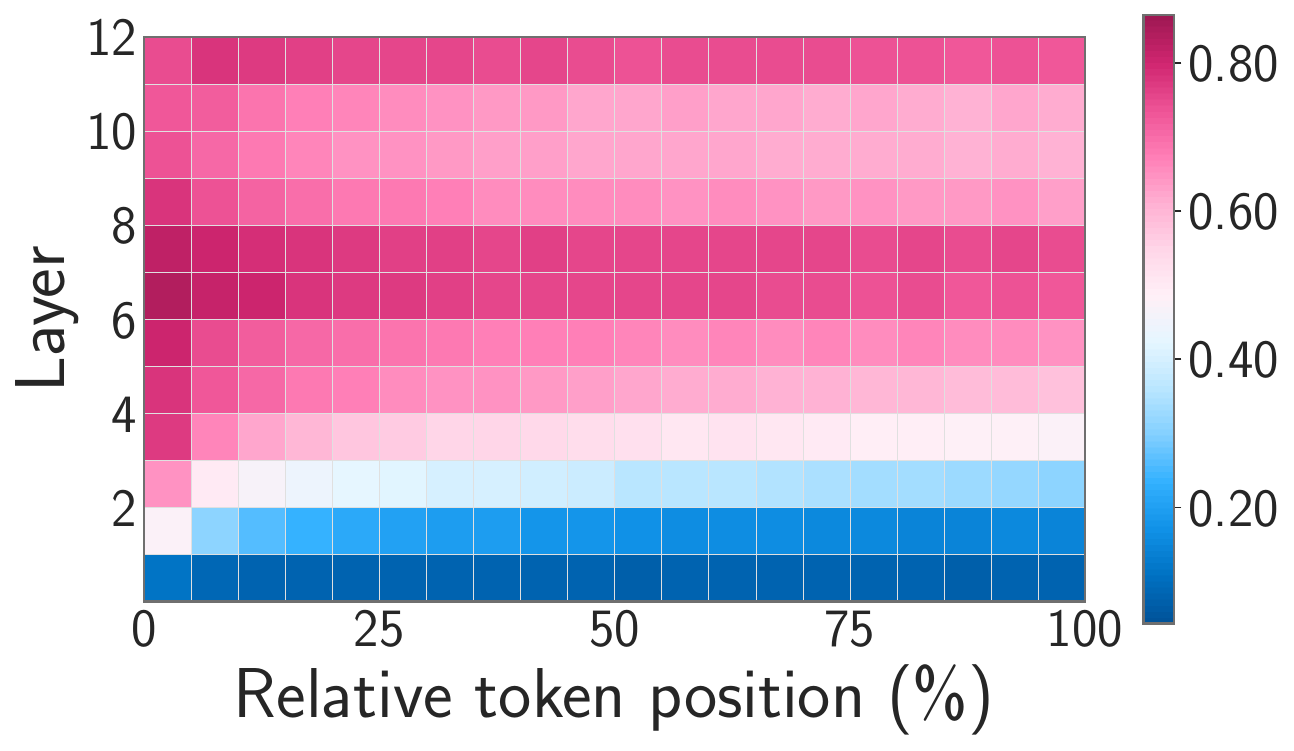}
    \includegraphics[width=\linewidth]{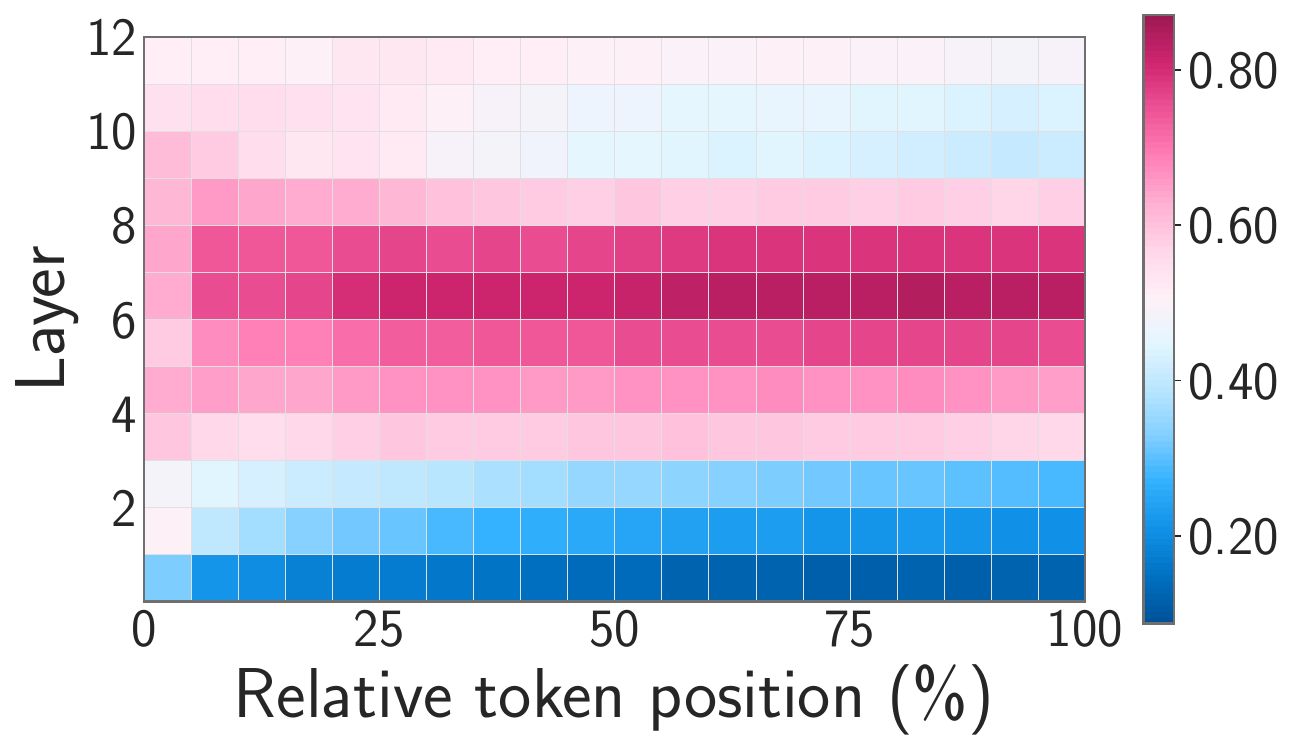}
    \caption{6 big layers}
    \label{fig:fig-combined-55router}
\end{subfigure}
\begin{subfigure}[t]{0.245\linewidth}
    \centering
    \includegraphics[width=\linewidth]{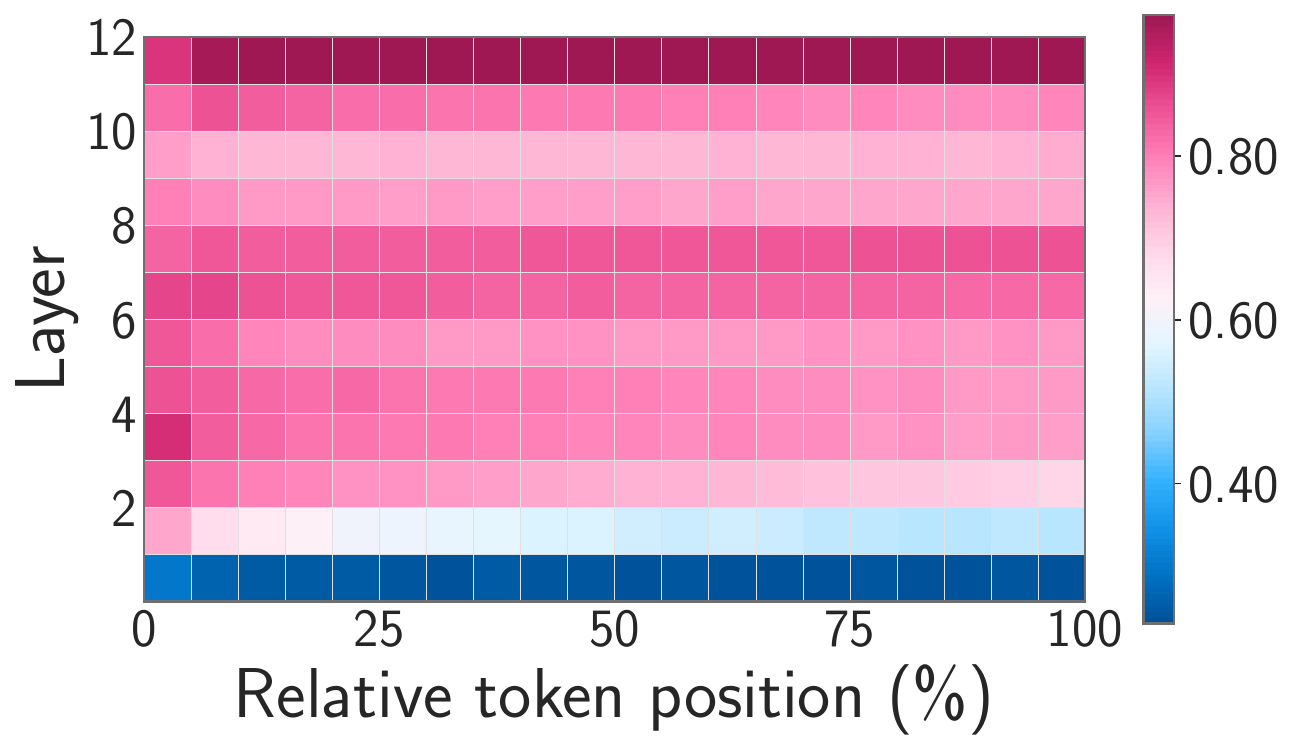}
    \includegraphics[width=\linewidth]{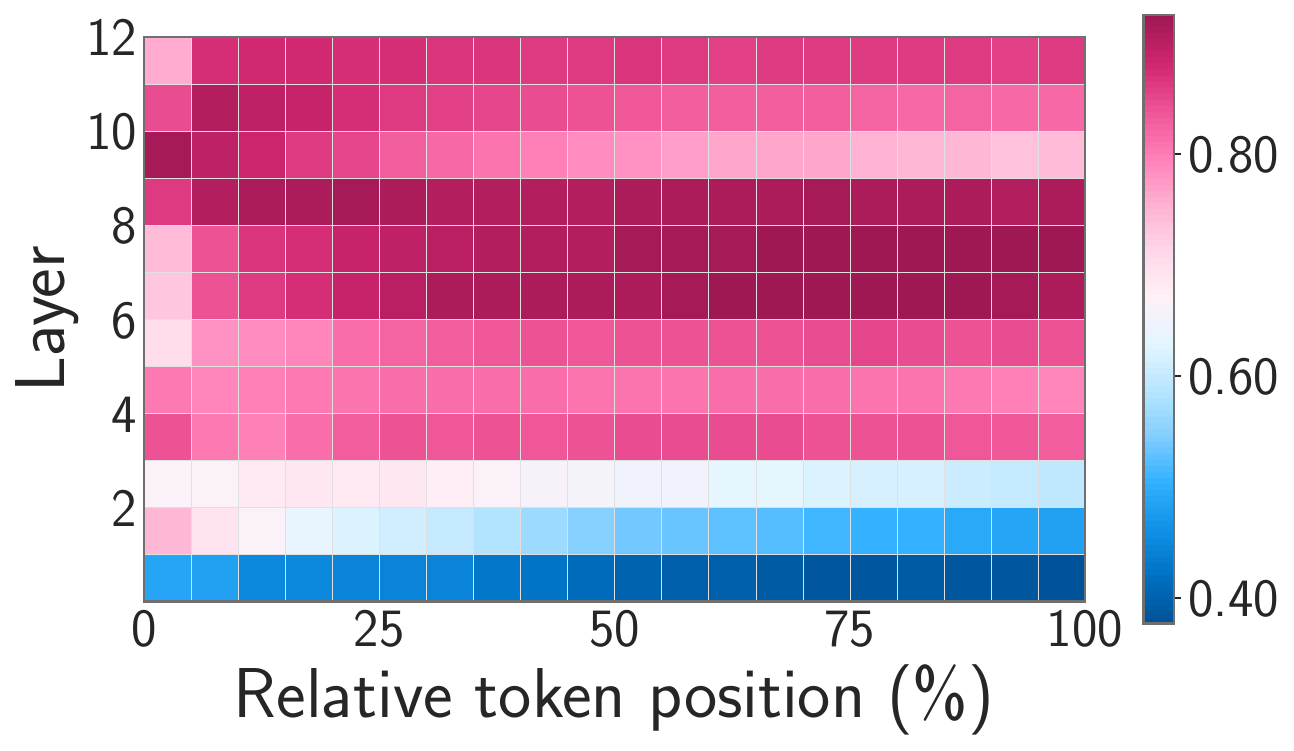}
    \caption{6 big layers}
    \label{fig:fig-combined-77router}
\end{subfigure}
\begin{subfigure}[t]{0.25\linewidth}
    \centering
    \includegraphics[width=\linewidth]{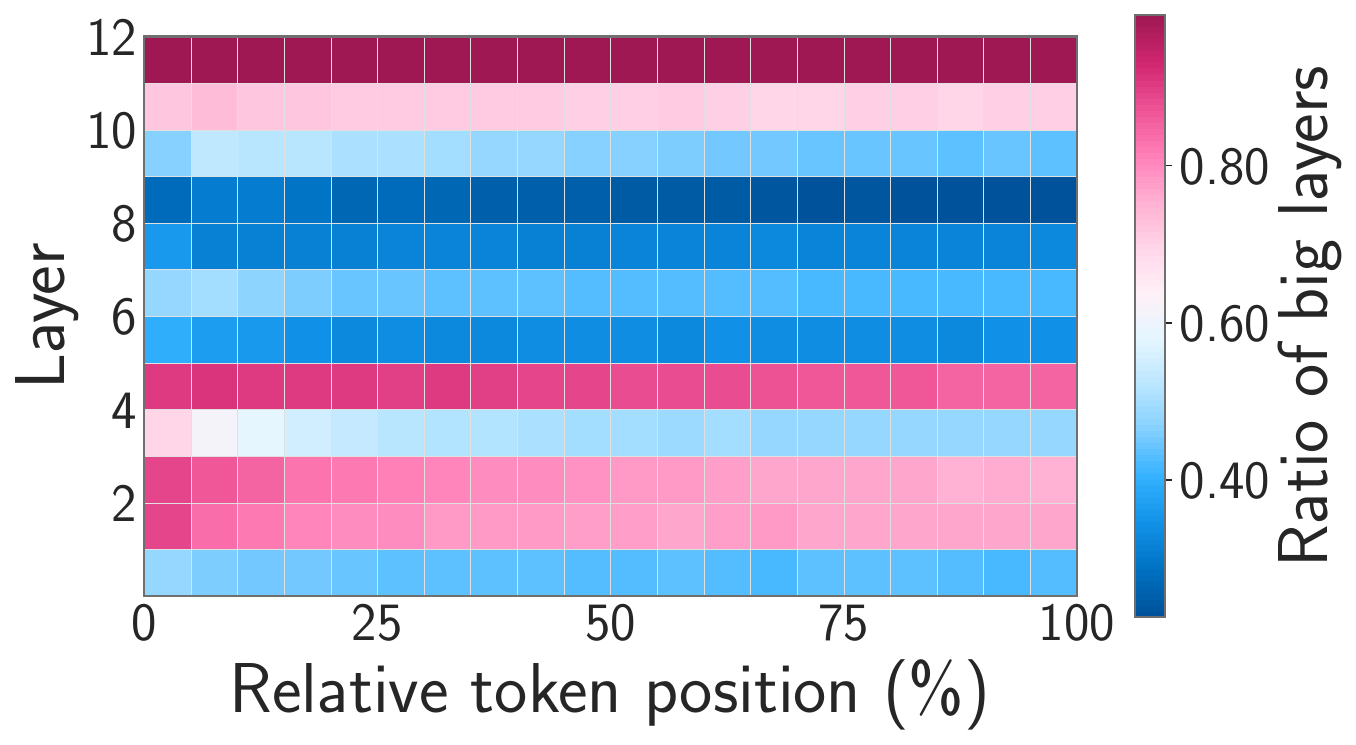}
    \includegraphics[width=\linewidth]{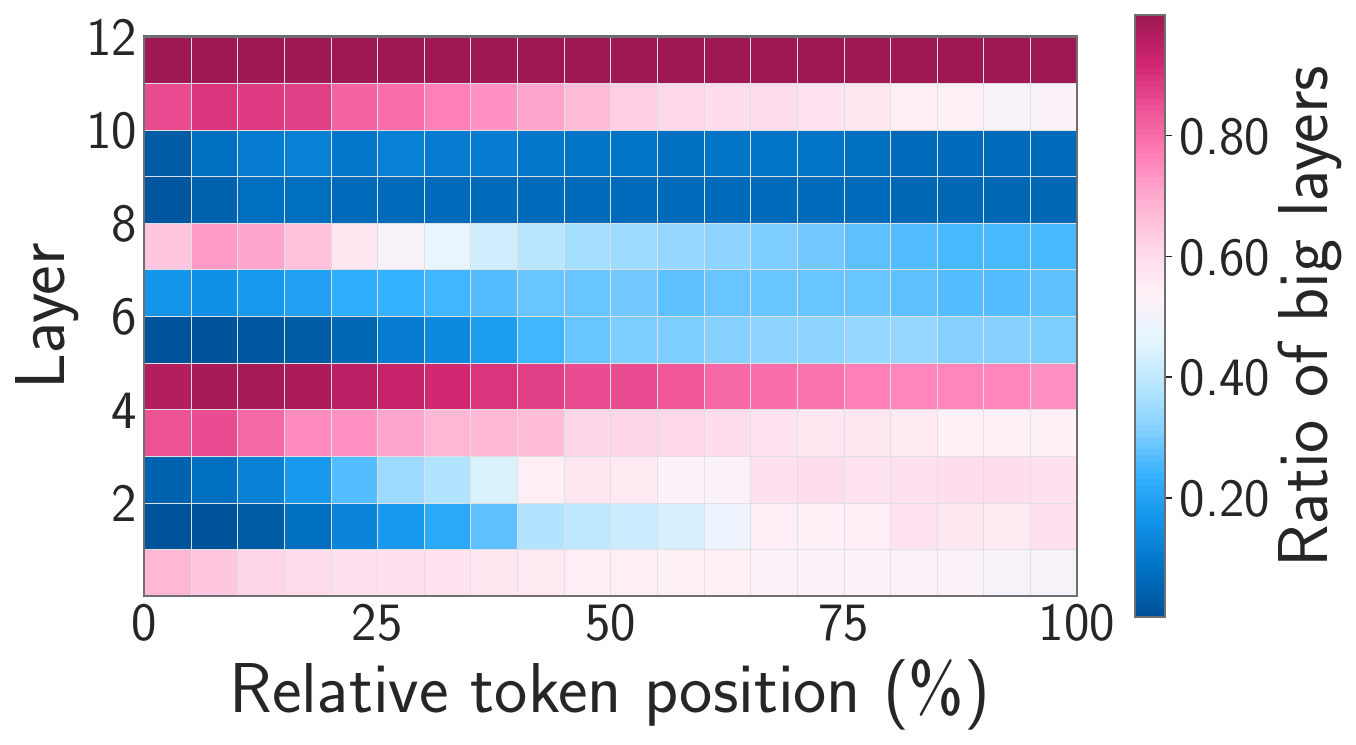}
    \caption{Skip layers}
    \label{fig:fig-combined-skiprouter}
\end{subfigure}
\caption{ 
Routing patterns for a learned router of conventional MoE on the C4 holdout set (top) and the code holdout set (bottom) under different budgets of big layers. (a) With a budget of average 4 big layers per token, later and middle layers are allocated big modules. Moreover, early tokens tend to need big moduls more often than later toekns. (b) With a budget of average 6 big layers, later layers chosen more frequently in contrast to oracle where uses big module more uniformly across layers. Again, early tokens tend to benefit from big modules more often than final tokens. (c) With a budget of average 8 big layers. (d) When 6 layers are skipped, the pattern is more irregular than oracle.
}
\vspace{-2mm}
\label{fig:router-analysis}
\end{figure*}

Despite advancements in adaptive methods for LLMs, the identification of optimal patterns for dynamic execution remains an unresolved challenge. This limitation hinders the ability to fully leverage the potential of these techniques. To overcome this obstacle, we propose Duo-LLMs demonstrated in Figure~\ref{fig:framework}, a novel framework designed for systematically studying adaptive computation in LLMs. Our approach integrates smaller auxiliary modules within each Feed-Forward Network (FFN) layer of the LLM, enabling dynamic token routing based on complexity. We introduce the concept of oracle routing, which exhaustively explores all possible routing options and selects the one that minimizes perplexity within a fixed computational budget. This method provides the theoretical optimal for per-token routing and, as we will demonstrate, offers valuable insights into LLM behavior and their use of computational resources during generation. For instance, as demonstrated in Figure \ref{fig:fig-combined-33}, given a small predefined computational budget (e.g. only 4 big layers out of 12) , the model tends to reserve larger modules for use in the later layers.
We compare this oracle-guided routing with the routing decisions made by a trained router similar to conventional Mixture of Expert (MoE) models shown in Figure \ref{fig:router-analysis}. The comparison highlights key differences in how these two approaches allocate computation, and underscores the limitations of current practical implementations in efficiently utilizing compute resources. These insights pave the way for improving adaptive computation in future models.

To summarize, here is our contribution built on the top of the proposed Duo-LLM framework: 
\begin{itemize}[leftmargin=*]
    \item  We introduce the use of oracles to identify optimal computation patterns, providing deeper insights into adaptive computation and routing mechanisms. 
   \item We demonstrate that oracle can identify patterns where activating only one big layer per token results in lower perplexity compared to using big layers across all layers. Generally, with the presence of oracle the quality of the optimal routing under a budget is proportional to the number of routing choices rather than the the number of big layers.
    \item We define the concept of token relative difficulty, capturing the potential benefit of additional computational resources beyond its loss value.
    \item We demonstrate the gap between theoretical optima and practical implementations by showing that trained routers often yield sub-optimal routing solutions compared to oracles.
\end{itemize}




\vspace{-2mm}
\section{Duo-LLM Framework}
\vspace{-2mm}

Our methodology consists of multiple stages designed to systematically investigate and optimize adaptive computation in LLMs. First, we train an LLM capable of selecting between a big and small FFN module within each layer. This choice enables adaptive computation, allowing the model to balance efficiency and performance. Next, we employ oracles to identify optimal routing patterns across different layers. In the final stage, we train routing mechanisms and compare their patterns to those identified by the oracles, highlighting discrepancies and potential areas for improvement.

\textbf{Stage 1: Training the Duo FFN Module.}
In the first stage, we focus on training a duo of a smaller and bigger FFN module. To achieve this, we put a big FFN and small FFN side by side within each layer. The attention part is shared among the two.


During this training process, we implement a random routing strategy to ensure that the smaller model can function interchangeably with the bigger model. Specifically, at the beginning of each FFN layer, tokens are routed to either the big or small module with a probability of 0.5. This means that, at each layer, tokens can either be processed by the big module or the small module. 

Alternatively, one could start from a pretrained bigger model, freeze it during Duo-LLM training and only finetune the small auxiliary modules. In the ablation section we have evaluated this strategy and found it performs worse compared to training the Duo model from scratch.  

It is worth noting that one could potentially train the smaller model using a learned routing strategy, similar to those employed in training of mixture of experts (MoE) models, where a router is trained alongside the experts. However, this approach can introduce bias in the training of the smaller modules, as the routing strategy itself influences the optimization process and training of the modules. Our methodology, in contrast, aims to train the smaller model in a more unbiased manner by relying on random routing during this initial stage.

{\bf Stage 2: Oracle-Guided Optimal Routing.}
Once we have independently trained the two interchangeable modules (big and small) per layer, we proceed to the second stage, where we determine the optimal routing strategy. To achieve this, we employ an oracle that exhaustively enumerates all possible routing paths and selects the one that results in the lowest perplexity on each token. 


Given  $n$ layers in the model, there are  $2^n$ possible routes when considering only the small vs. big routing options, and  $3^n$ routes when including the skip option. However, since our primary goal is to study the efficiency-accuracy trade-off in adaptive computation, for some of our experiments we constrain our analysis to a subset of routing strategies that adhere to a fixed computational budget. For example, we might limit feasible routes to those that utilize the big module 30\% of the time for each token. This allows us to identify the optimal routing strategy under a specified computation budget. 
We apply this oracle-guided routing on sequences from a randomly sampled holdout set, and the results of this analysis are presented in the following section.

{\bf Stage 3: Approximating Oracle Routing with Learned Strategies.}
While oracle-guided routing provides valuable insights, it is not feasible in real-world applications due to its exhaustive nature. Therefore, in the third stage of our methodology, we explore the feasibility of approximating the oracle's optimal routing with a learned routing strategy.

We implement a router training algorithm similar to the conventional mixture of experts (MoE) strategy \citep{shazeer2017outrageously, fedus2021switch, zoph2022stmoe, wang2024hmoe}, where the router learns to assign tokens to the appropriate module (small or big) based on input complexity. In these settings, the total loss is a combination of the cross-entropy loss $\mathcal{L}_{CE}$ and a weighted load balancing loss. The load balancing loss ensures equal usage of each expert within each layer. However, instead of enforcing load balancing per layer, we use a budget loss that enforces a computational budget across all layers for each sequence. Our total loss is defined as follows:
$$\mathcal{L}_{tot} = \mathcal{L}_{CE} + \alpha \mathcal{L}_{budget}.$$
Our router includes a learnable linear layer $W_{r,l}$ for each layer $l$, which outputs the logits $h_{l}$. The probabilities assigned to each module are given by:
$$P_{l,x} = \frac{e^{h_{l,x}}/\tau}{\sum_{j} e^{h_{l,j}}/\tau},$$
where $x$ represents either the big or small FFN. The output of the layer is calculated as $\sum_{x} P_{l,x} \cdot H_{l,x}$, where $H_{l,x}$ denotes the hidden states obtained by routing layer $l$ through module $x$. The temperature parameter $\tau$ is gradually increased to enforce hard assignment.
The budget loss is defined as $$\mathcal{L}_{budget} = (\frac{\sum_{i=1}^{L}{P_{i,\text{big}}}}{L} - \text{budget})^2,$$ which enforces a specified budget for the use of big layers across the entire sequence. By applying this budget loss over all layers, rather than using a per-layer load balancing loss, the model gains flexibility to allocate more computational resources to certain layers, facilitating the discovery of optimal routing patterns.

We then evaluate how closely the learned strategy approximates the oracle's routing decisions and assess its performance in terms of both efficiency and accuracy, as well as its routing and decoding dynamics. By comparing the learned routing strategy to the oracle's optimal routing, we aim to evaluate the feasibility of deploying such strategies in real-world scenarios with limited computational resources. Previous work, such as Megablocks \citep{gale2022megablocks}, has demonstrated that block-sparse matrix multiplication with experts can be efficiently executed on a single GPU. Given that the auxiliary models in Duo-LLM are small enough to coexist with the larger model on a single node, we hypothesize that any routing strategy that reduces FLOPs will also reduce latency. However, developing an efficient implementation is beyond the scope of this work.

\vspace{-2mm}
\section{Experiments}
\vspace{-2mm}
In this section, we begin by presenting our experimental setup, we then provide an analysis of the results obtained from the oracle focusing on identifying optimal routing patterns. The concept of relative token difficulty will be introduced next and the findings from our experiments conducted on the holdout sets are shared. Finally, we analyze the router's behavior, highlighting the key differences in performance compared to the oracle.

\vspace{-2mm}
\subsection{Setup}
\vspace{-2mm}
We trained our Duo-LLM model using a 12-layer architecture, with each layer resembling those in Llama 2~\citep{touvron2023llama}. The model has a hidden dimension of 2560 and incorporates two feed-forward networks (FFNs): a bigger FFN with an inner dimension of 10,240, and a smaller one that is 16 times smaller, with an inner dimension of 640. Both FFNs share the same attention mechanism. The model has a total of 1.399 billion parameters, distributed as follows: 59 million from the smaller FFNs, 944 million from the bigger FFNs, 314 million from the attention, and the remainder from embeddings. During stage one of Duo-LLM training, tokens were randomly assigned to either FFN at each layer with a 50\% probability. We trained the model on 300 billion tokens from FineWeb~\citep{penedo2024fineweb}, Wiki and Flan datasets (from Dolma~\citep{soldaini2024dolmaopencorpustrillion}), and Python code (Stack-v2)~\citep{lozhkov2024starcoder}. Additional details regarding the performance during training can be found in Section \ref{sec:ablations}.

We then determined the optimal routing using oracles on two holdout sets: one consisting of 1,024 subsamples from the C4 validation set~\citep{DBLP:journals/corr/abs-1910-10683}, and the other comprising 1,024 Python code samples sourced from GitHub repositories with MIT licenses~\citep{huggingface_codeparrot}. 
A budget per token was specified for oracle to meet the hypothetical efficiency constraints, like in an application on a device the model can handle only 50\% of big module. 
The router was trained on 10 billion tokens from the same datasets as the original training set, with the budget loss to match the resource constraints.

\begin{figure*}
\centering
\begin{subfigure}[t]{0.3\linewidth}
    \centering
    \includegraphics[width=\linewidth]{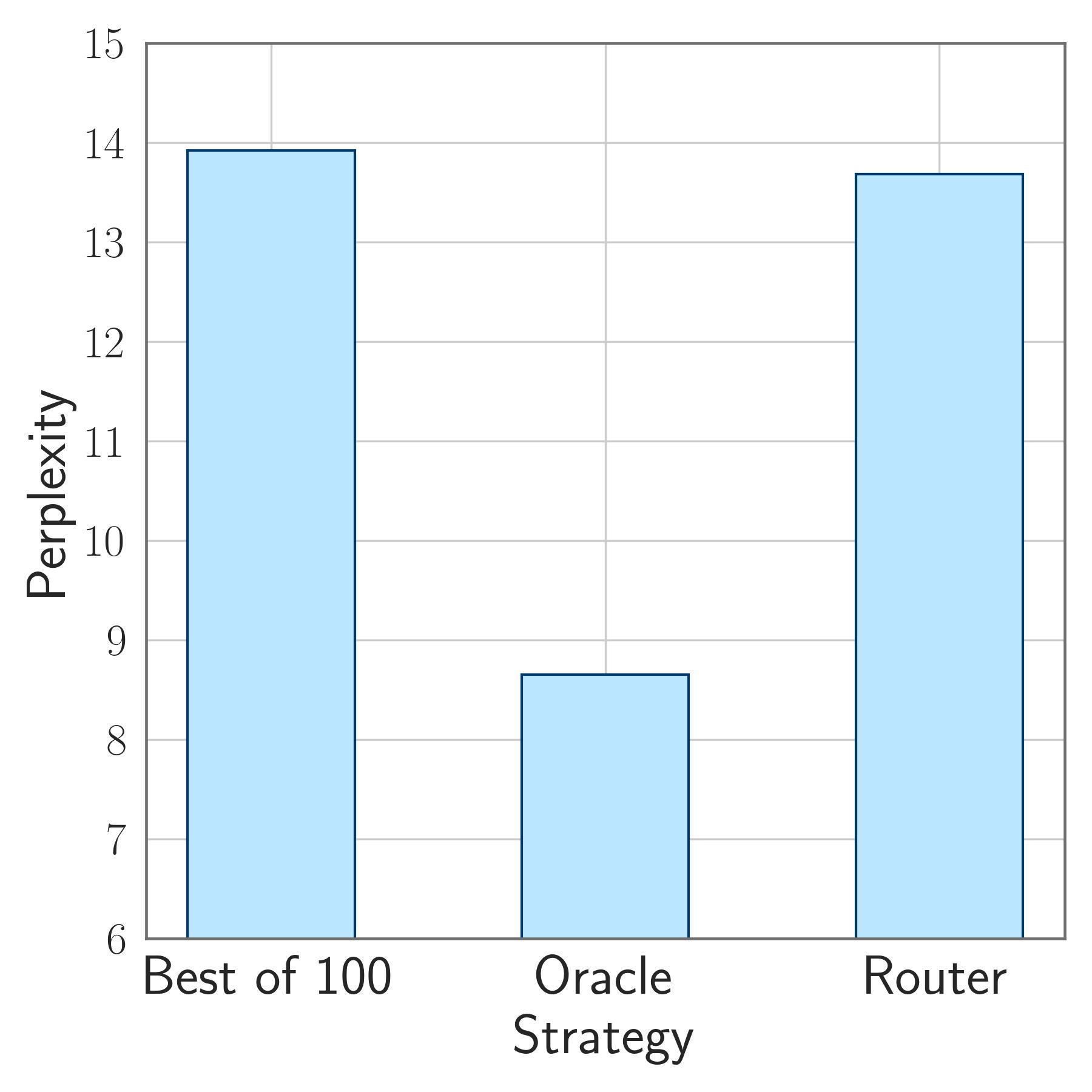}
    \caption{Best strategy}
    \label{fig:fig-oracle-vs-100}
\end{subfigure}
\begin{subfigure}[t]{0.39\linewidth}
    \centering
    \includegraphics[width=\linewidth]{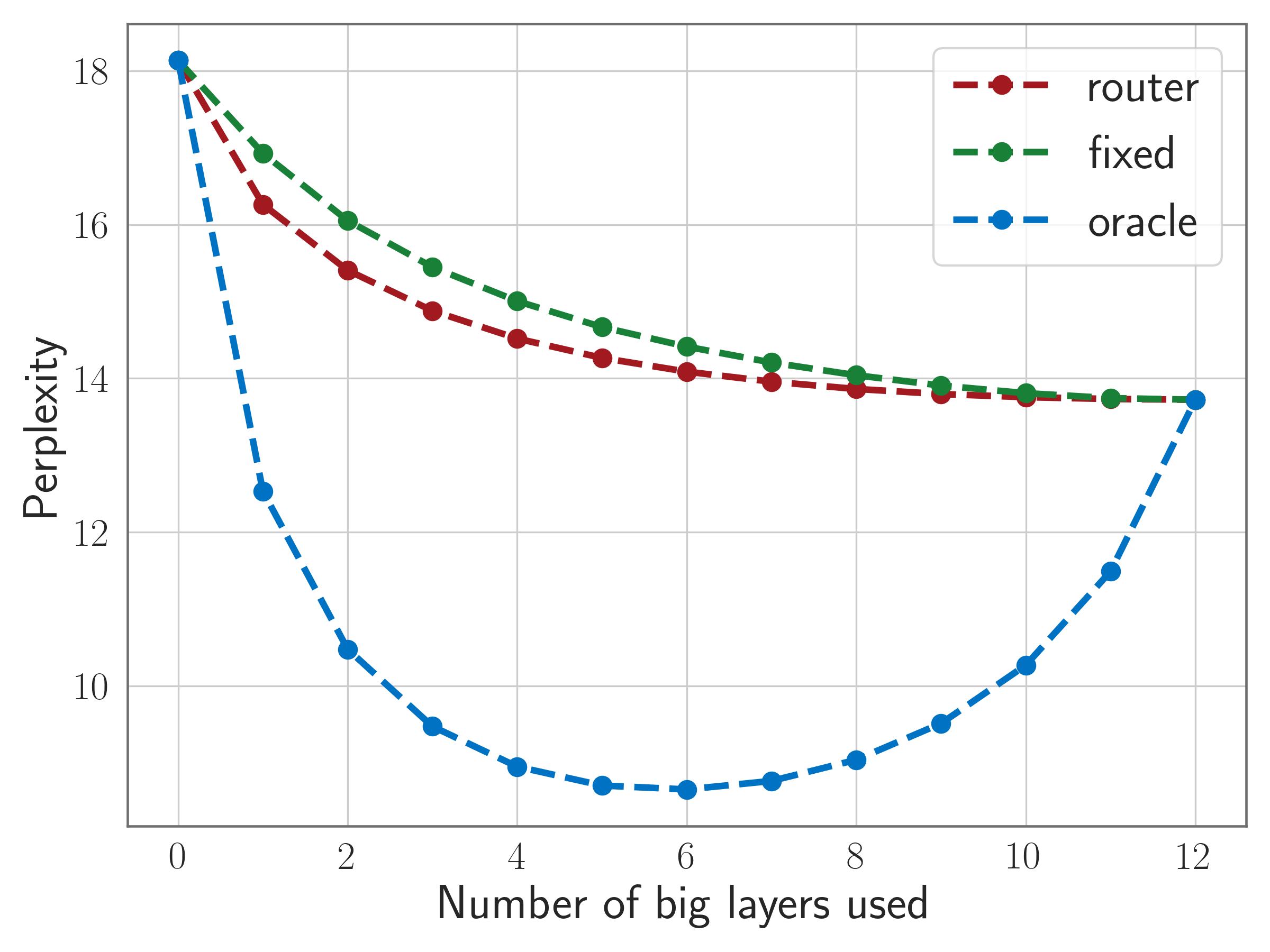}
    \caption{Flexible vs fixed routing}
    \label{fig:fig-flexible-fixed}
\end{subfigure}\hfill
\begin{subfigure}[t]{0.3\linewidth}
    \centering
    \includegraphics[width=\linewidth]{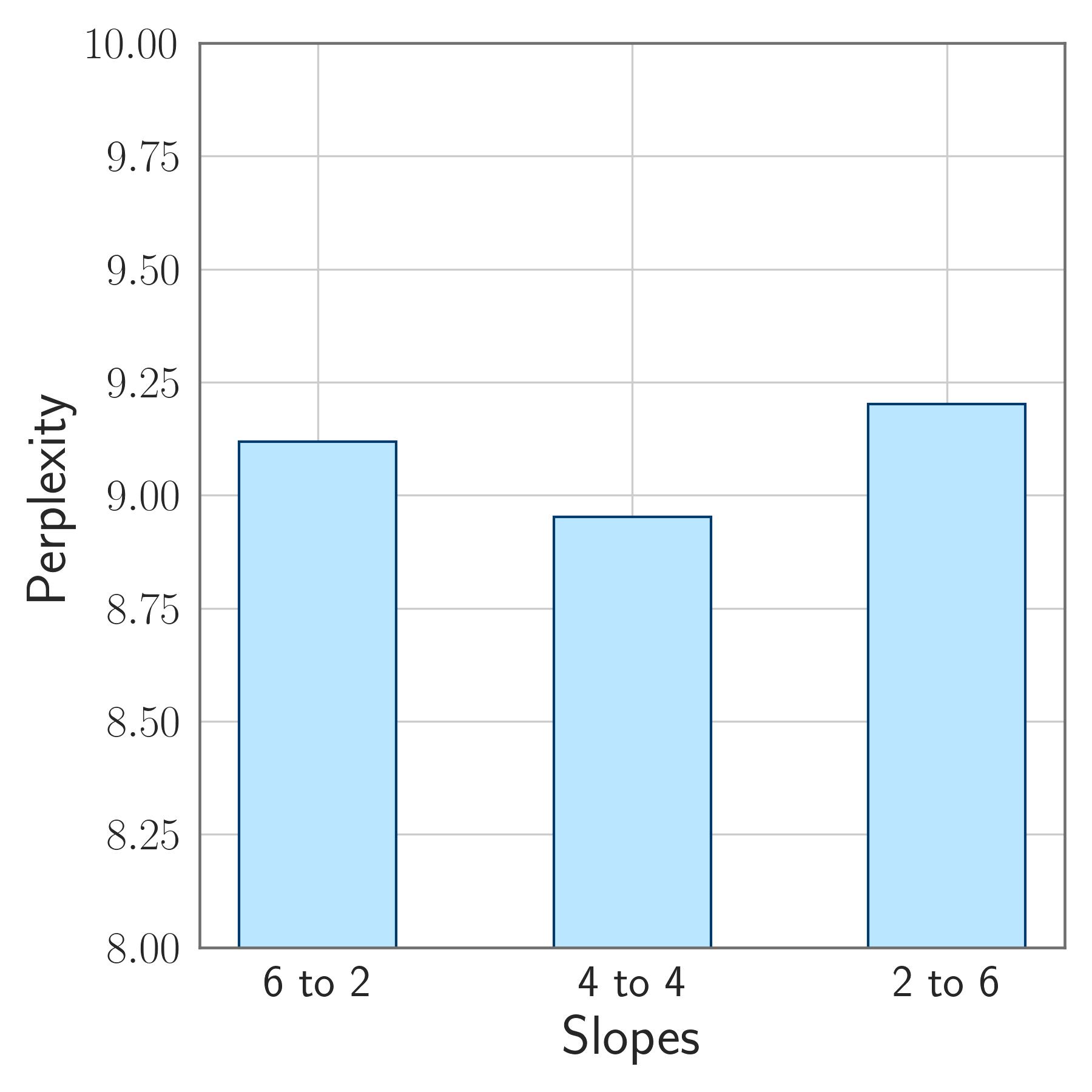}
    \caption{Slope comparison}
    \label{fig:fig-slope-ppx}
\end{subfigure}
\caption{ Oracle perplexity results. (a) The oracle surpasses random routing patterns, even when selecting the best out of 100 random trials; the router's performance is closer to fixed patterns than to the oracle. (b) The oracle achieves the lowest perplexity using 6 big layers per token, outperforming fixed patterns and routers. Notably, using only one big layer, the oracle attains a lower loss than when all layers are big.  (c) The oracle prefers a consistent budget per token: gradually reducing the budget from 6 to 2 big layers increases loss compared to consistently using 4 big layers throughout. }
\label{fig-main}
\end{figure*}

\vspace{-2mm}
\subsection{Results} 
\vspace{-2mm}
In this section, we first examine the properties of the oracle and its routing behavior. We then introduce the concept of token difficulty, which leverages the presence of both small and big modules within the LLM. Finally, we explore the characteristics of a learned router based on conventional MoE training and compare its routing patterns with those of the oracle.

\subsubsection{Oracle Analysis} 
\textbf{Optimality of the Oracle.}
To demonstrate the effectiveness of our oracle, we compare its perplexity on the C4 holdout set against an alternative approach that evaluates 100 different random patterns and selects the best one. As shown in Figure \ref{fig:fig-oracle-vs-100}, the oracle achieves a significantly lower perplexity than the random pattern approach. This confirms that the optimal patterns identified by the oracle are significantly better than random ones, affirming the oracle's effectiveness in discovering optimal routing strategies.

Next we consider a scenario where only the small and big modules are available as routing options, without the possibility of skipping layers. Specifically, we explore how varying the budget allocated to the use of big modules impacts the final perplexity. In Figure~\ref{fig:fig-flexible-fixed}, we compare the perplexity on the C4 holdout set when using the oracle, the trained router, or the best fixed layer pattern across different budgets of big layer usage. The \emph{oracle} optimizes the routing pattern on a per-token basis, whereas the the so called \emph{fixed} strategy applies the same routing pattern to all tokens, computes the average loss over tokens, and selects the optimal configuration accordingly. The name \emph{router} refers to a router trained using conventional MoE training. 

Firstly, the oracle consistently achieves much lower perplexity across all budget levels. Secondly, both the \emph{fixed} and \emph{router} approaches show improvements in perplexity as more big layers are employed. However, the oracle reveals a particularly intriguing pattern: perplexity decreases until 6 big layers are used, then begins to rise as the number increases to 12.
Thirdly, oracle's solution with only one big layer  achieves lower perplexity than the configuration where all layers use the big module.
Utilizing 6 big layers out of 12 results in the lowest perplexity is perhaps due to the increased number of possible configurations (ie $^{12}C_6$) , which enhances the probability of selecting better-performing options. These findings demonstrate that incorporating a small auxiliary FFN can significantly improve model accuracy under optimal routing. Bridging this performance gap between the router training in conventional MoEs and the oracle remains as an interesting direction for future research.


\textbf{Routing Patterns of Oracle.}
Furthermore, we analyzed the pattern of selecting big modules across layers. Figure~\ref{fig:fig-combined-33} illustrates how the oracle allocates layers for module selection under the constraint that only 4 out of 12 big modules can be chosen. Consistent patterns emerged in both the C4 dataset and the coding holdout set. Interestingly, the oracle tends to allocate more small modules to the earlier layers, conserving the computational budget by reserving the bigger modules for the later layers in the FFN. Figure~\ref{fig:fig-combined-55} shows the case where the budget is evenly distributed between small and big layers, with 6 out of 12 big modules available for selection. The oracle assigns similar proportions across all layers. However, when the budget for big layers increases to 8 out of 12 the oracle predominantly assigns the bigger modules to the initial layers as shown in Figure~\ref{fig:fig-combined-77}. This pattern is noteworthy, as it suggests that the final layers need to meet a certain threshold for optimal performance. Once this threshold is satisfied, the earlier layers benefit from additional computational resources. If the threshold is not met, prioritizing the final layers appears to be the more advantageous strategy.
We then analyzed the usage pattern in the presence of skipping. For simplicity, we focused on the case where the oracle either selects a big layer or skips it. In Figure~\ref{fig:fig-combined-skip}, the oracle made some layers to be mostly skipped throughout the sequence, while others are predominantly retained. No clear pattern of early exits \cite{elhoushi2024layerskip} or skipping of early layers \cite{corro2023skipdecode} was observed. This suggests that existing skipping methods in the literature may have room for further improvement.

\begin{figure}[t]
    \centering
    \includegraphics[width=1\linewidth]{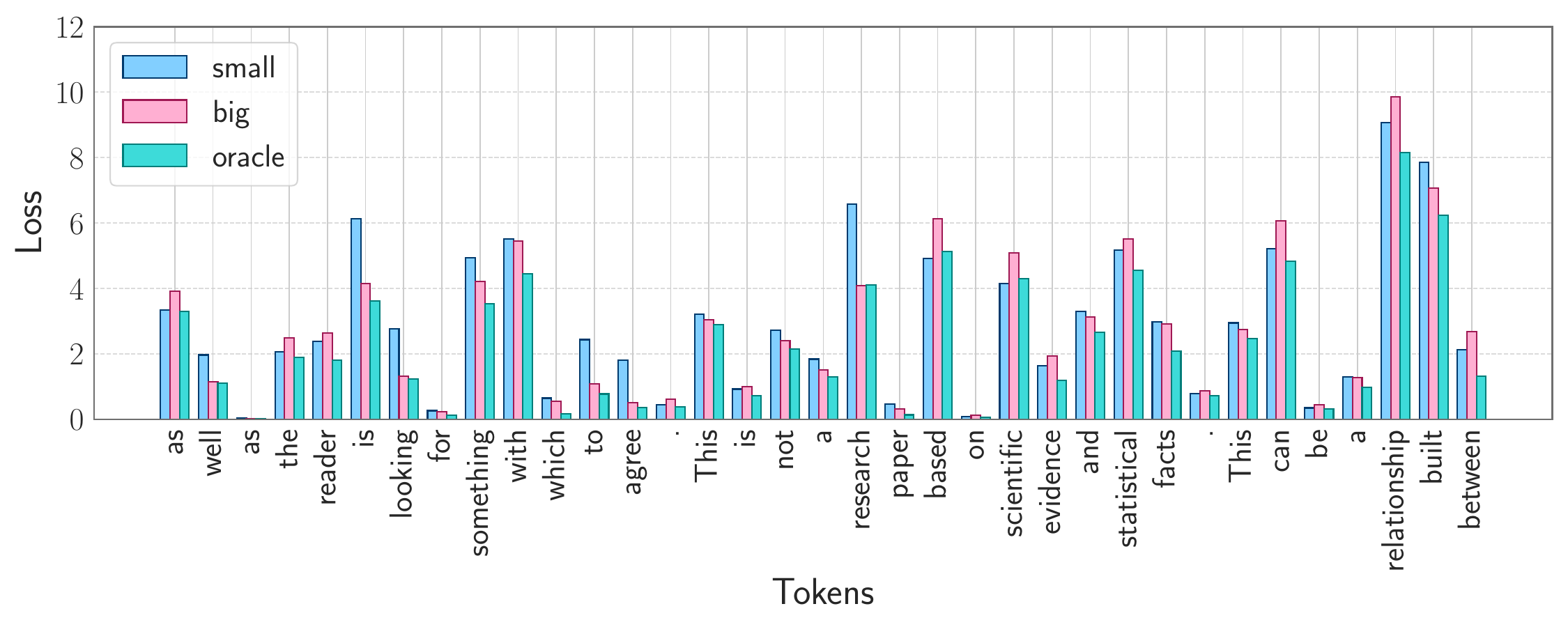}
    \caption{The loss for each token varies depending on whether the small, big, or oracle model is used. Some tokens, especially those following a clause, are inherently unpredictable due to the many possible continuations. For example, the phrase \emph{``This can be a''} could be followed by various words, making the choice of ``\emph{relationship}'' uncertain, and leading to high loss, even with increased compute. In contrast, some tokens are more predictable based on context. For instance, the word ``\emph{research}'' following the phrase ``\emph{This is not a}'' can be inferred from the surrounding context, such as when the text is part of an analysis.
    }
   \label{fig:small-vs-big-vs-oracle-loss}
\end{figure}

\begin{figure*}[t]
  \centering

  \begin{subfigure}[t]{0.49\linewidth}
    \centering
    \includegraphics[width=\linewidth]{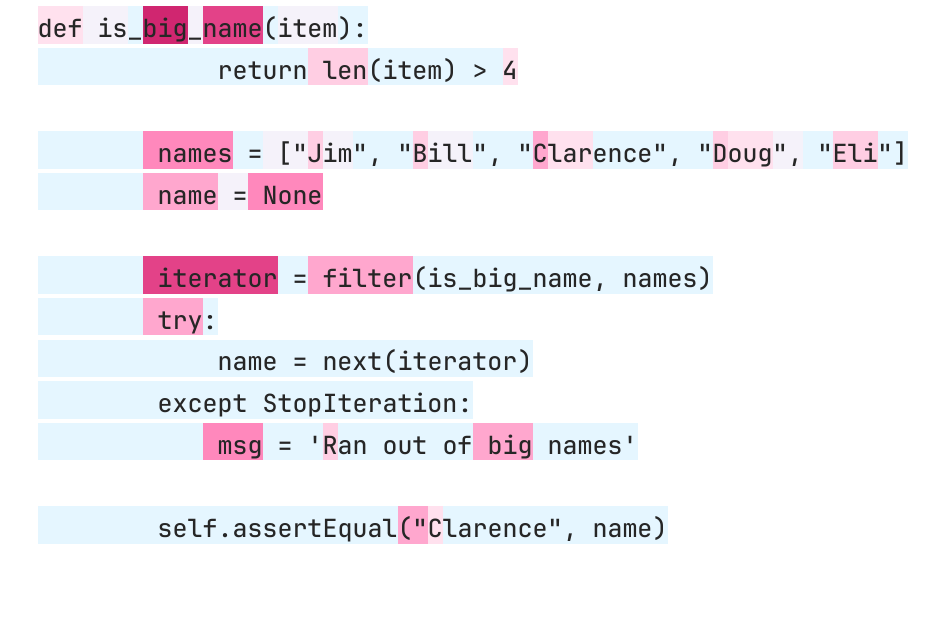}
    \caption{Difficult tokens when using small modules}
    \label{fig:small-code}
\end{subfigure}
\begin{subfigure}[t]{0.49\linewidth}
    \centering
    \includegraphics[width=\linewidth]{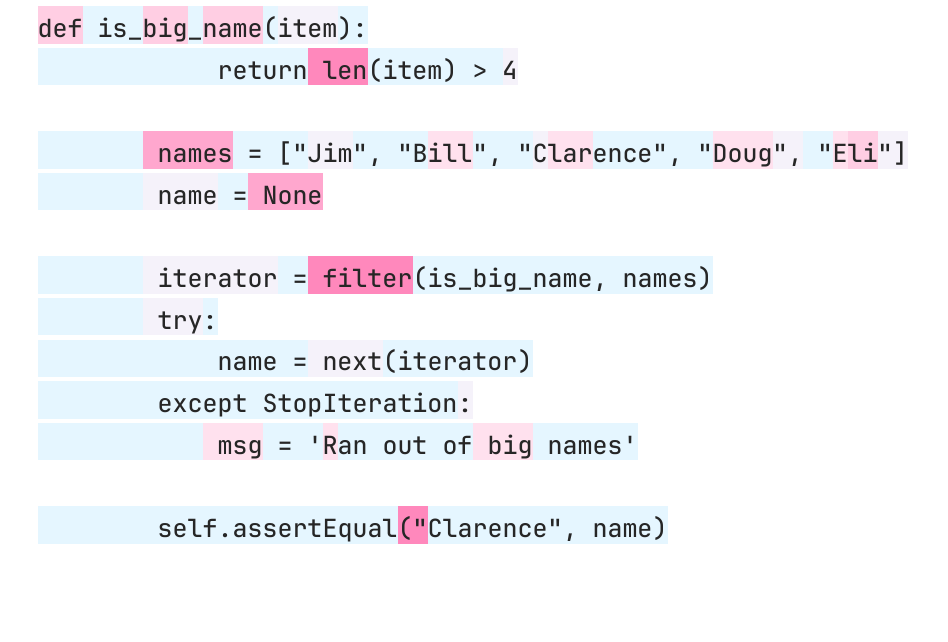}
    \caption{Tokens with highest loss gap }
    \label{fig:gap-oracle-small-code}
\end{subfigure}
\caption{Token difficulty in a Python code snippet. Blue represents lower difficulty values, while red indicates higher values. (a) when difficulty is measure using  the loss from small model alone, tokens at the beginning of line tend to be more challenging, which is intuitive. (b) when measuring the loss gap between the oracle and the small model, the  highest values correspond to tokens that can benefit from additional computation. In this case, some tokens at the beginning of the line, e.g. \texttt{try} or \texttt{iterator}, are no longer as important since they cannot be further improved, likely due to their inherent uncertainty given the context. Instead, tokens like \texttt{len} rank higher in terms of benefiting from more compute compared to when only the small model’s loss is considered.}

\end{figure*}

\subsubsection{Token Difficulty} 
Previous research suggests that tokens with higher loss values are considered difficult and require routing through larger layers \citep{ethayarajh2021understanding, salehi2023sharcs}. 

\textbf{C4 Dataset.}
To validate the conventional notion of token difficulty, we compared loss values across three configurations: using small modules for all layers, using large modules for all layers, and an optimal oracle solution. Our findings indicate that for certain high-loss tokens, switching from the small model to the large one, or even using the oracle, does not significantly improve performance, as shown in Figure~\ref{fig:small-vs-big-vs-oracle-loss}. For example, it is clear in the figure that the word ``\emph{research}'' achieves a lower loss when a bigger model or oracle is used, likely because its meaning can be inferred from context. These are tokens for which additional compute make a difference.
However, some tokens, such as ``\emph{relationship}'' following the phrase ``\emph{This can be a}'', present many possible continuations and are inherently difficult. In such cases, assigning more capacity does not provide a notable benefit. This observation suggests that relying solely on loss values may not be the most reliable indicator of token difficulty or the need for additional computational resources. 

\textbf{Coding Dataset.} Extending the above observation to the coding domain, in Figure~\ref{fig:small-code}, we observe that the first tokens of each line of code are considered challenging for smaller models. However, in Figure~\ref{fig:gap-oracle-small-code}, only the tokens \texttt{filter}, \texttt{len}, \texttt{names}, and \texttt{None} achieve a significant reduction in loss when using the oracle. Based on these results, we suggest that the potential for loss reduction, rather than purely token difficulty, may be considered for adaptive routing. Further exploration of this novel yet practical concept of token difficulty presents a promising avenue for gaining mechanistic insights into adaptive computation and, more broadly, the decoding dynamics of LLMs.

\subsection{Router Performance Analysis} 

\textbf{Sub-optimality of the Learned Router.}
We now examine the performance of a router which is trained using the conventional MoE router training scheme, and compare it to the oracle. Figure~\ref{fig:fig-oracle-vs-100} demonstrates that the router's performance aligns more closely with the best of 100 trials rather than the oracle. This significant performance gap highlights the potential benefit of training a surrogate model, akin to the approach outlined in~\citep{cai2024flextron}.

\textbf{Routing Patterns of Learned Router.}
Figure~\ref{fig:router-analysis} illustrates how router performs on C4 and code holdout set.
In Figure~\ref{fig:fig-combined-33router}, ~\ref{fig:fig-combined-55router}, ~\ref{fig:fig-combined-77router} where 4, 6 and 8 layers, respectively, are allocated as big layers, the router tends to assign more compute to later layers. This contrasts with the oracle, which only allocates more budget to the later layers when the budget is limited (only 4 big layers of big modules).  Figure \ref{fig:fig-combined-skiprouter} shows the case when skipping layers is an option, the router favors skipping in the later layers, resembling an early exit strategy. This behavior suggests that the router fails to discover the more complex patterns identified by the oracle, which could yield better perplexity outcomes. 

\begin{figure}[tbp]
    \centering
    \begin{minipage}{1\textwidth}
        \centering
        \includegraphics[width=\linewidth]{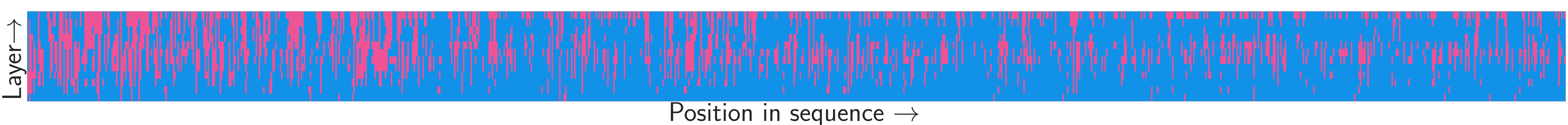}
        \label{fig:sub1}
    \end{minipage}
\vspace{-2mm}
    \begin{minipage}{1\textwidth}
        \centering
        \includegraphics[width=\linewidth]{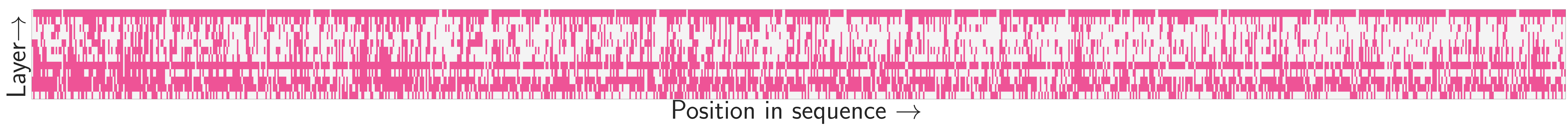}
        \label{fig:sub2}
    \end{minipage}
\vspace{-2mm}
    \begin{minipage}{1\textwidth}
        \centering
        \includegraphics[width=\linewidth]{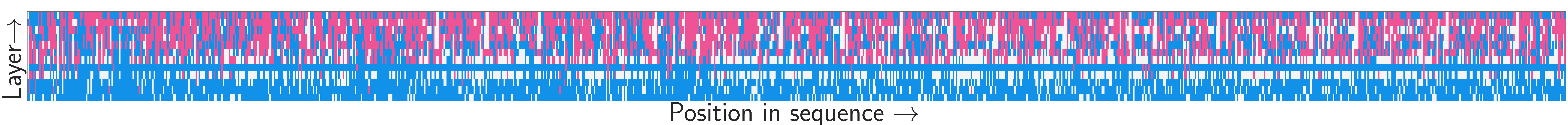}
        \label{fig:sub3}
    \end{minipage}
    \caption{Routing patterns of a learned router on a single sequence: blue indicates the use of the small module, red represents the use of the big module, and white denotes a skip. (Top) The model uses 70\% small modules and 30\% big modules. (Middle) The model uses 50\% big modules and 50\% skips. (Bottom) The model uses 50\% big modules, 30\% small modules, and 20\% skips.}
    \vspace{-2mm}
    \label{fig:router}
\end{figure}

\textbf{Individual Routing Behavior.}
The figures above show the aggregate behavior of the learned router. To better understand how a conventional MoE router behaves, we examine its decisions for a single randomly selected sequence from the C4 dataset. 
Figure~\ref{fig:router} illustrates how its  operation under three different settings. Notably, the router increasingly opts to skip or use the smaller module toward the end of the sequence, a behavior that differs from the oracle’s. 
 Figure~\ref{fig:fig-slope-ppx}, when the oracle is instructed to start with 6 big layers per token and reduce to 2 layers towards the end (or vice versa), the perplexity is worse compared to a constant 4-layer budget throughout the sequence. This discrepancy between the oracle and router patterns highlights the challenge of training routers for imbalanced mixtures.
 Similar patterns were observed in the OPT-1.3B model's router, as detailed in Appendix \ref{appendix:opt}.

\subsection{Ablations}
\label{sec:ablations}
In this section, we present the results of pretraining Duo-LLMs compared to traditional dense model training, and examine how pretraining accuracy varies with different training methods.

\begin{figure}[tbp]
    \centering    
    \begin{minipage}{0.34\textwidth}
        \centering
        \subcaptionbox{Arc Easy \label{fig:arc-easy}}
        {
        \includegraphics[width=\linewidth]{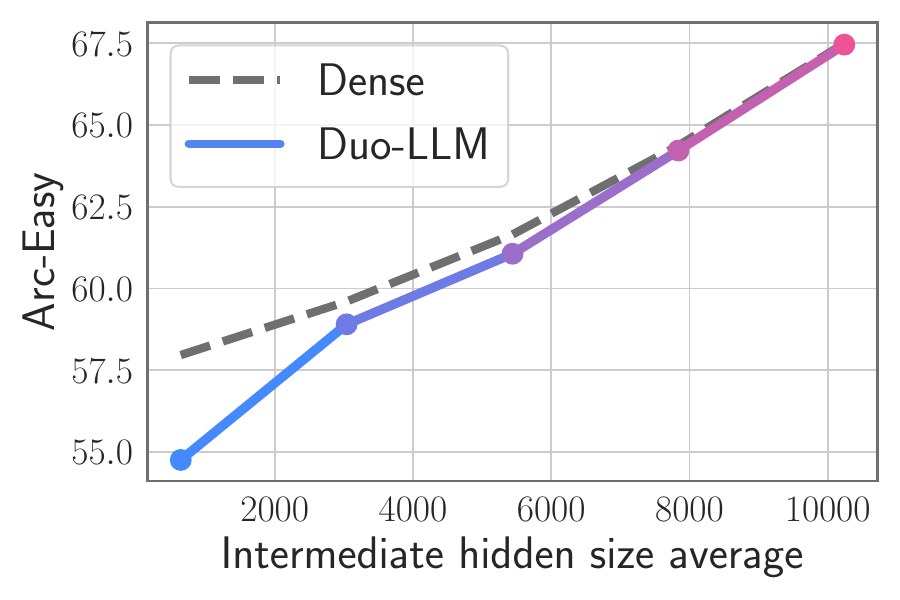}}                
    \end{minipage}
    \begin{minipage}{0.34\textwidth}
        \centering
        \subcaptionbox{Hellaswag        \label{fig:hellaswag}}{
        \includegraphics[width=\linewidth]{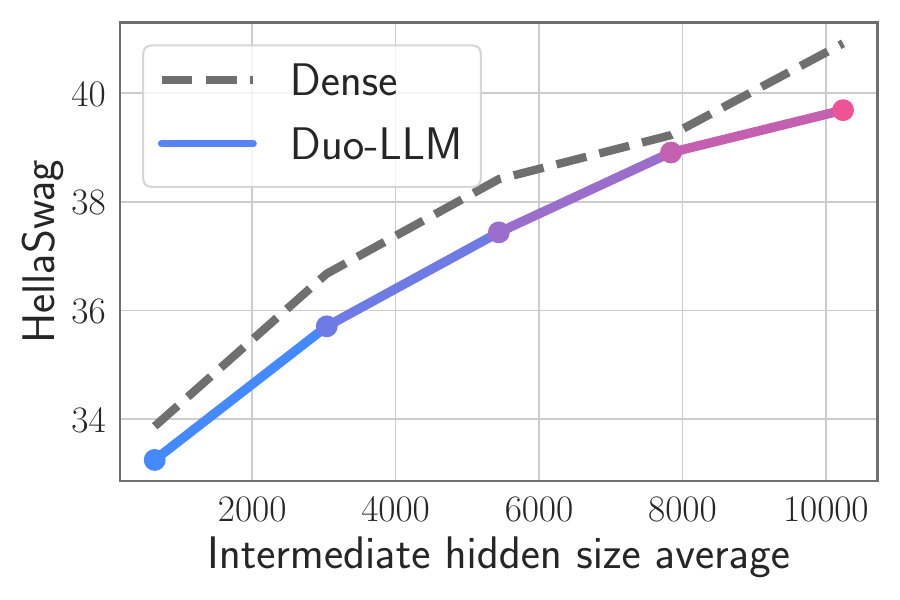}}        
    \end{minipage}
    \begin{minipage}{0.28\textwidth}
    \centering
    \subcaptionbox{Dou-LLM training strategies 
    \label{tab:training-method}}{
        \begin{tabular}{@{}lS[table-format=2.2]S[table-format=2.2]@{}}
    \toprule
     & \multicolumn{2}{c}{Dataset} \\
     \cmidrule(l){2-3}
    {\makecell{Training \\ Method}} & {\makecell{Arc \\ Easy}} & {\makecell{Hella \\ Swag}} \\
    \midrule
    From scratch & 61.06 & 37.44 \\
    Frozen big   & 53.35 & 33.63 \\
    \bottomrule
\end{tabular}
    }
\end{minipage}

    \caption{Accuracy of stage 1 training: (a) We compare Duo-LLMs trained in stage 1 with dense models. We set different probabilities for selecting big modules in Duo-LLMs and train equal sized dense counterparts. The arc easy of Duo-LLM gets closer to dense counter part when big layers is mostly selected (b) The hellaswag accuracy of dense models are higher than Duo-LLMs under different sizes (c) Training big modules and freezing them then finetuning small modules alongside big modules is much less accurate than training both from scratch.}
    \label{fig:table_fig}
\end{figure}

\textbf{Comparison with Dense Training.} To assess the accuracy of our Duo-LLM pretrained model, we compared it with dense models, varying the probability of big module across [0, 0.25, 0.5, 0.75, 1.0]. We also trained dense models with equivalent FLOPs under a similar training regime using 300B tokens. At a probability of 0, only the small modules are used, while at 1, only the big modules are utilized. The results shown in Figures \ref{fig:arc-easy} and \ref{fig:hellaswag}, demonstrate that our model approaches the performance of the dense model when predominantly big modules are utilized and exhibits lower accuracy when only the small modules are used. Notably, the big module in the Duo-LLM has processes only half of the tokens, as they are distributed between the two modules. Despite this, it still achieves performance nearly on par with traditional dense model training. Further research into pushing the Pareto-optimal boundaries of the accuracy-efficiency curve is encouraged.

\textbf{Duo-LLM Training Method.} In addition to training from scratch, we evaluated the accuracy of Duo-LLMs under an alternative setting, where the big module was trained on 200B tokens, then frozen, followed by training the small modules alongside it on an additional 100B tokens. The total number of tokens processed in both training methods is the same. As shown in Table~\ref{tab:training-method} there is a significant accuracy gap between the two approaches, leading us to recommend training Duo-LLMs from scratch for optimal performance.


\section{Related Works}

\textbf{Adaptive Computation.}
Adaptive computation has been extensively studied within the research community \citep{lin2019conditional, cai2019onceforall, puigcerver2020scalable}. More recently, it gained increased attention because of LLMs' huge computational demands. Matformer~\citep{devvrit2023matformernestedtransformerelastic} trains one model and smaller subsets of it can be selected for efficient deployment. Mixture of Depths~\citep{raposo2024mixtureofdepths} dynamically skips layers of transformer, and Flextron~\citep{cai2024flextron} converts a trained model to a nested elastic structure where computation dynamically happens. Here, we study how adaptive computation works to provide insights on the underlying routing patterns.

\textbf{Mixture of Experts.}
Mixture of Experts (MoE) architectures achieve efficient scaling by adapting to inputs and distributing tasks across multiple specialized, smaller networks \citep{fedus2021switch, shazeer2017outrageously, zoph2022stmoe}. Recently, architectures incorporating heterogeneous experts have been proposed \citep{wang2024hmoe, zhou2022mixtureofexperts}. While our work is similar to these approaches, since Duo-LLMs also employ a heterogeneous mixture of experts, we utilize this architecture as a framework to study adaptive computation, rather than as a standalone architecture.

\textbf{LLM Functional Analysis.} A substantial body of research has sought to elucidate the inner workings of neural NLP models. Numerous studies have examined neuron behavior, investigating what individual neurons or groups of neurons represent~\citep{NEURIPS2020_92650b2e, li2022lazy, durrani2020analyzing}, as well as the internal features of LLMs \citep{wendler2024llamas}. In contrast, our work analyzes behavior at the module level. While much of the existing research has concentrated on the role of self-attention \citep{voita-etal-2019-analyzing, Clark_2019, vig-belinkov-2019-analyzing}, and some on the function of Feed-Forward Networks (FFNs) \citep{geva2020transformer} or both components \citep{yu2023joma}, we focus on studying the mechanisms of adaptive computation and mixture of experts.

\section{Conclusion}
The results from our experiments highlight the effectiveness of the Duo-LLM framework in understanding adaptive computation dynamics in LLMs. We demonstrated that the oracle can select configurations with significantly higher precision than random choices. Notably, the oracle identified that activating only a single big layer per token can achieve lower perplexity compared to using large layers throughout all layers. Additionally, our findings reveal that later layers have a certain capacity that, once fulfilled, shifts the prominence to earlier layers. We also observed that layers which skip in the earlier parts of a sequence tend to benefit from further skipping later on.

Moreover, we showed that token difficulty is a relative metric, and in some cases, allocating higher compute budgets to certain tokens results in wasted resources. Lastly, we highlighted the gap between the performance of the trained router and the oracle, demonstrating how their routing patterns differ and suggesting that a surrogate model could be beneficial in future work. These findings confirm that our proposed approach offers a robust framework for studying adaptive computation, with the potential for further refinement through more advanced routing strategies in future research.

We acknowledge that our framework has practical limitations. The oracle requires exhaustive iteration over all possible options to find the minimum loss, which is computationally demanding and time-intensive. Additionally, computing the oracle’s loss requires access to ground truth labels, making it impractical for real-world deployment. As a result, Duo-LLM is primarily positioned as a theoretical framework for studying adaptive computation rather than a practical heterogeneous MoE setup for efficient adaptive computation. Future research is needed to develop surrogate metrics that eliminate the need for ground truth loss and to devise methods that bypass the exhaustive enumeration of all options. These advancements would not only impact adaptive computation but also provide deeper insights into the decoding dynamics of transformer-based Large Language Models.

\textbf{Acknowledgement.} The authors would like to thank Kumari Nishu, Antonie Lin, Mohammad Samragh, Max Horton, Devi Krishna, Fartash Faghri, Oncel Tuzel, Hadi Pouransari, Arsalan Farooq, Mohammad Rastegari, Qi Shan, Ghong Wang, and Devang Naik for valuable discussions.

\bibliography{references}
\bibliographystyle{plainnat}

\appendix

\appendix
\section{Learned Router in OPT Model}
\label{appendix:opt}
\begin{figure*}[ht]
\centering
\begin{subfigure}[t]{0.99\linewidth}
    \centering
    \includegraphics[width=\linewidth]{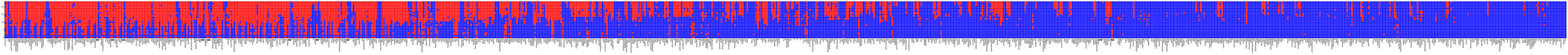}
    \caption{Two options: big module (red) and small module (blue)}
    \label{fig:opt-1}
\end{subfigure}

\begin{subfigure}[t]{0.99\linewidth}
    \centering
    \includegraphics[width=\linewidth]{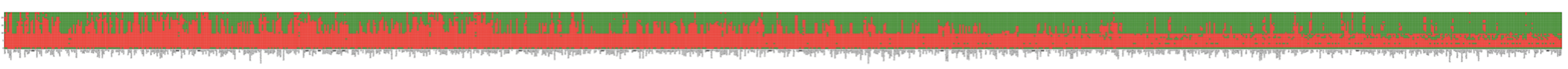}
    \caption{Two options: skip (green) and big module (red)}
    \label{fig:opt-2}
    
\end{subfigure}
\begin{subfigure}[t]{0.99\linewidth}
    \centering
    \includegraphics[width=\linewidth]{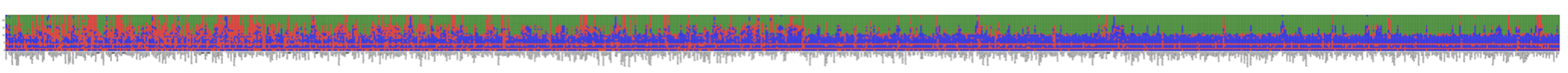}
    \caption{Three options: big module (red), small module (blue), skip (green)}
    \label{fig:opt-3}
\end{subfigure}

\vspace{-2mm}
\caption{ 
Learned routing of Duo-LLM with varying budget and options (a) when 70\% small and 30\% big modules is used. (b) when 50\% small and 50\% skip is used. (c) when 30\% big,  20\% skip, and 50\% small is used.
}
\label{fig-opt-results}
\end{figure*}

We conducted an additional study by training Duo-LLMs on the OPT 1.3B model with the same setup as described in the main paper, and trained a router on top of it. Figure~\ref{fig-opt-results} illustrates the routing patterns across various budgets and settings. The key observations from these three setups are as follows:
\begin{itemize}[leftmargin=*]
\item Early layers tend to use the small module, allowing later layers to benefit from additional compute.
\item  Early tokens often utilize the big modules, while later tokens, having sufficient context built up, require less compute for decoding.
\item  When skipping is an option, early layers and tokens tend to avoid skipping too much, whereas later layers and tokens tend to skip more, as sufficient compute has already been spent earlier in the sequence or model.
\end{itemize}

\end{document}